\documentclass{article} 
\usepackage{main_style,times}

\usepackage{amsmath,amsfonts,bm}

\def\eqref#1{equation~\ref{#1}}

\def\1{\bm{1}}

\def\mA{{\bm{A}}}

\def\mC{{\bm{C}}}

\def\mK{{\bm{K}}}

\def\mM{{\bm{M}}}

\def\mQ{{\bm{Q}}}

\def\mV{{\bm{V}}}
\def\mW{{\bm{W}}}
\def\mX{{\bm{X}}}

\DeclareMathAlphabet{\mathsfit}{\encodingdefault}{\sfdefault}{m}{sl}
\SetMathAlphabet{\mathsfit}{bold}{\encodingdefault}{\sfdefault}{bx}{n}

\newcommand{\R}{\mathbb{R}}

\usepackage{xcolor}
\usepackage{colortbl}
\usepackage{algorithm}
\usepackage{booktabs}
\usepackage{multirow}
\usepackage{listings}
\usepackage{caption}
\usepackage{array}
\usepackage{algpseudocode}
\usepackage{hyperref}
\usepackage{url}
\usepackage{graphicx}
\usepackage{wrapfig}
\usepackage{subcaption}
\usepackage{tabularx}
\usepackage{threeparttable}
\usepackage{enumitem}
\usepackage{svg}
\usepackage{calc}
\usepackage{tikz}
\usepackage{csquotes}
\newcommand*\circled[1]{\tikz[baseline=(char.base)]{%
            \node[shape=circle,fill=black,draw,inner sep=0.1pt, font=\small,text=white] (char) {#1};}}
\usepackage{enumitem}
\usetikzlibrary{shapes.arrows}
\newcommand{\FancyUpArrow}{\begin{tikzpicture}[baseline=-0.3em]
\node[single arrow,draw,rotate=90,single arrow head extend=0.2em,inner
ysep=0.2em,transform shape,line width=0.05em,top color=green,bottom color=green!50!black] (X){};
\end{tikzpicture}}
\newcommand{\FancyDownArrow}{\begin{tikzpicture}[baseline=-0.3em]
\node[single arrow,draw,rotate=270,single arrow head extend=0.2em,inner
ysep=0.2em,transform shape,line width=0.05em,top color=red,bottom color=red!50!black] (X){};
\end{tikzpicture}}
\DeclareMathOperator{\mode}{mode}
\DeclareMathOperator{\myargmax}{argmax}
\DeclareMathOperator{\myargmin}{argmin}
\definecolor{forestgreen}{HTML}{228B22}
\definecolor{wrongred}{HTML}{d66344}
\title{NoVo: Norm Voting off Hallucinations with Attention Heads in Large Language Models}

\author{Zheng Yi Ho \And Siyuan Liang \And Sen Zhang \And Yibing Zhan \And Dacheng Tao
}

\finalcopy 
\begin{document}

\maketitle

\begin{abstract}
Hallucinations in Large Language Models (LLMs) remain a major obstacle, particularly in high-stakes applications where factual accuracy is critical. While representation editing and reading methods have made strides in reducing hallucinations, their heavy reliance on specialised tools and training on in-domain samples, makes them difficult to scale and prone to overfitting. This limits their accuracy gains and generalizability  to diverse datasets. This paper presents a lightweight method, Norm Voting (NoVo), which harnesses the untapped potential of attention head norms to dramatically enhance factual accuracy in zero-shot multiple-choice questions (MCQs). NoVo begins by automatically selecting truth-correlated head norms with an efficient, inference-only algorithm using only 30 random samples, allowing NoVo to effortlessly scale to diverse datasets. Afterwards, selected head norms are employed in a simple voting algorithm, which yields significant gains in prediction accuracy. On TruthfulQA MC1, NoVo surpasses the current state-of-the-art and all previous methods by an astounding margin---at least 19 accuracy points. NoVo demonstrates exceptional generalization to 20 diverse datasets, with significant gains in over 90\% of them, far exceeding all current representation editing and reading methods. NoVo also reveals promising gains to finetuning strategies and building textual adversarial defence. NoVo's effectiveness with head norms opens new frontiers in LLM interpretability, robustness and reliability.
\end{abstract}

\section{Introduction}
\label{sec:intro}
One of the most significant challenges facing Large Language Models (LLMs) today is their tendency to hallucinate---outputs that are factually incorrect or entirely fabricated \citep{hallu_survey1}. This flaw is particularly serious in high-stakes applications like finance and healthcare, where even small errors can lead to huge losses and compromised patient safety \citep{finance_hallu,med_hallu_harm}. Reducing factual hallucinations is a critical research area with major practical benefits, essential for realising the full potential of LLMs to revolutionise these industries by enhancing efficiency and decision-making, and safeguarding against costly and harmful errors \citep{hallu_causes_harm}.

Given these serious risks and the high cost of retraining LLMs, it is crucial to find affordable techniques to reduce factual hallucinations. Although inference techniques such as retrieval augmentation and prompt engineering work well, they come with significant limitations: latency and external dependencies, and the need for user expertise, respectively \citep{rag_survey, prompt_engineering_survey}. In response, we turn to representation editing and reading methods (REAR) \citep{repe}, which operate within the model, ensuring rapid response times and eliminating the need for external data or user interaction. REAR methods reduce hallucinations by modifying or extracting factual information encoded in LLMs' latent feature vectors (hidden states), such as attention heads \citep{hidden-states-factual-rs}. This process often requires specialized tools such as probes and autoencoders \citep{iti,truthx}, trained and tuned on in-domain samples. Thus, existing REAR methods are difficult to scale and prone to overfitting, leading to limited accuracy gains and generalizability to diverse datasets. Tackling these limitations is crucial, since REAR methods can improve factuality with minimal costs, latency, and user friction; highly desirable attributes for practical applications.

\begin{figure}[h]
    \centering
    \includegraphics[width=\textwidth]{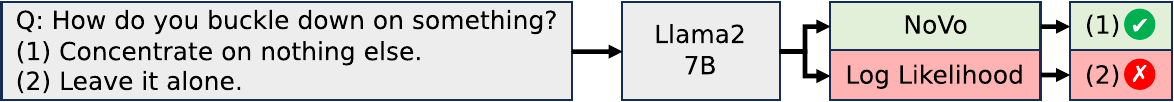}
    \caption{\textbf{Overview of our method}. NoVo improves factuality in diverse MCQs.}
    \label{fig:overview}
\end{figure}

This paper presents \textbf{No}rm \textbf{Vo}ting (NoVo), a more accurate and generalizable REAR method for reducing factual hallucinations in LLMs. Following previous works, we evaluate NoVo on multiple-choice questions (MCQs), which are excellent tests of factuality and serve as an important foundation for more complex tasks. Figure \ref{fig:overview} shows an overview of this process. NoVo achieves higher MCQ accuracies by efficiently measuring \emph{latent truth} \citep{repe} in certain attention head norms, thus avoiding the log likelihood layer, which can induce hallucinations by favouring fluency over factuality \citep{mitigation-weakness1}. NoVo first selects attention head \emph{\textbf{norms}} that correlate with truth, using only inference passes over 30 random samples, allowing NoVo to scale to numerous datasets. Then, selected head norms participate in majority \emph{\textbf{voting}}, which acts as an ensemble of weak learners \citep{strength-of-weak-learners}, resulting in more accurate predictions. Since scalability is crucial for practical use, NoVo is made lightweight by design, not requiring any specialised tools or training, such that it can be evaluated across a diverse range of reasoning, factuality, and understanding datasets. As far as we know, we are the first to explore attention head norms' potential as a measure of latent truth. This raises exciting questions about their wider roles for mechanistic interpretability in LLMs, and how they can be used to address hallucinations.  

On TruthfulQA MC1, an unsolved hallucination benchmark, NoVo achieves a new state-of-the-art (SOTA) accuracy of 78.09\% on a 7B model, substantially outperforming the log likelihood and the best REAR method by at least 24 and 19 accuracy points. NoVo scales and generalizes well to 20 diverse datasets featuring varied topics and formats, with significant gains in over 90\% of them, dramatically surpassing previous REAR methods, which could only be evaluated on a few factuality benchmarks. Promising results on Adversarial GLUE and DeBERTa finetuning suggests that NoVo also enhances robustness against textual attacks and finetuning accuracy. We analyse NoVo's inner workings and find that head norms primarily measure truth in both fine-grained token relationships and overall sequence structure. This insight could aid future interpretability research. Ensembling head norms will improve accuracy, especially if their error variability is high. Additionally, we find that NoVo performs well in avoiding misleading assumptions, but may struggle when scenarios require stereotyping. We hope NoVo's detailed analyses and strong results encourages practical use.

Our main contributions can be summarised in three points: \circled{1} We introduce NoVo, a more scalable, accurate, and generalizable REAR method. It uses attention head norms to improve factual accuracy. \circled{2} NoVo achieves a SOTA accuracy of 78.09\% on TruthfulQA MC1, significantly outperforming all previous methods by a huge margin---at least 19 accuracy points. NoVo scales and generalizes well to 20 diverse datasets, which is unprecedented amongst current REAR methods. \circled{3} We provide actionable insights on using NoVo effectively, analysing its novel approach and use of head norms to measure truth. This explanation could aid in future research for interpretability and reliability.

\section{Related Works}
\label{sec:related_works}
\textbf{Representation Editing}\quad Some REAR methods involve manually modifying hidden states during inference towards hidden state clusters, formed by the forward propagation of true and false sequences \citep{ccs_burns}, to improve factual accuracy. All methods here require cross-fold training on in-domain samples from the test set, with some set aside for validation. Inference Time Intervention (ITI) edits specific attention head hidden states towards those clusters \citep{iti}, using custom-built linear probes and visualisation tools. Similarly, TruthForest (TrFr) edits heads toward multiple directions \citep{trfr}, while Truthx edits concepts of truth disentangled from hidden states with a deep autoenconder \citep{truthx}, as a specialised tool. 

\textbf{Representation Reading}\quad There are decoding strategies that use hidden states to improve the factuality of LLMs without editing. This includes Decoding by Contrasting Layers (DoLa) \citep{dola}, which extracts factual information in intermediate layers, tuned with in-domain samples, and Induce-then-Contrast Decoding (ICD) \citep{icd}, which contrasts LLM outputs with a special hallucinatory model trained on a large external dataset. RePE \citep{repe} uses a different, more direct approach with highly curated templates and samples, to measure truth in hidden states, with an special technique known as linear artificial tomography. All these methods here extract factual information from hidden states without editing, to improve factual accuracy. 

Unlike current REAR methods, NoVo uses only attention head norms and does not require any external modules, custom-built probes, special techniques, in-domain sample training, or curated resources. This makes NoVo lightweight, allowing to scale and generalize to numerous datasets. Together with a simple voting algorithm, NoVo is also significantly more accurate.

\section{Method}
\label{sec:method}
\subsection{Background}
\label{subsec:setup}
\textbf{Preliminary}\quad  \citet{ccs_burns} found that some attention heads were linearly separable to classify true or false statements, which formed the basis of representation editing. However, \citet{iti} found that orthogonal hyperplanes could also classify well above chance, indicating the multifaceted nature of truth. REAR methods typically define a claim as true if supported by publicly available and reliable evidence, following \citet{truthfulqa}. Yet, the latter also framed truth as a probability measure. We observe that most existing REAR methods operated on the basis that truth is binary, yet at the same time hinted at its complex nature. We believe that clarifying this inconsistent interpretation is crucial, by acknowledging that the truthfulness of a sequence is multifaceted and non-discrete. This leads us to look for a broad and continuous measure of truth, denoted $T$.

\textbf{Setup}\quad In the forward pass of an auto-regressive decoder transformer LLM, token sequences of length \(s\) are embedded and featurized through multiple layers as hidden states, before reaching the likelihood layer for next-token prediction. Each layer includes a multi-head attention (MHA) module and a two-layer fully-connected perceptron (FFN). The likelihood and embedding layers are not counted. An LLM with \(L\) layers and \(H\) heads per MHA will have a total of \(LH\) heads throughout the network. The MHA at layer $\displaystyle l \in \{1, 2 \dots, L \}$ takes as input $\displaystyle \mX^{(l-1)}\in\R^{s\times d}$ from the previous layer and projects each feature in the sequence to their key, query and value states
\begin{align}
   \displaystyle \mQ^{l} &=\displaystyle \mX^{(l-1)}\displaystyle \mW_{query}^{l} &
   \displaystyle \mK^{l}&=\displaystyle \mX^{(l-1)}\displaystyle \mW_{key}^{l}  &
   \displaystyle \mV^{l}&=\displaystyle \mX^{(l-1)}\displaystyle \mW_{value}^{l}
\end{align}
ignoring the bias term, where $\displaystyle \mQ^{l},\mK^{l},\mV^{l} \in \R^{s \times d}$ and \(d\) is the model dimension. Splitting them on the column axis gives  $\displaystyle \mQ^{l,h},\mK^{l,h},\mV^{l,h} \in \R^{s \times d^\prime}$ for $\displaystyle h \in \{1, 2 \dots, H \}$ and $d^\prime=d/H$. The context vectors, or attention heads, $\displaystyle \mC^{l,h} \in \R^{s \times d^\prime}$, is thus computed via the attention mechanism as
\begin{align}
\label{eqn:attn_mechanism}
    \displaystyle \mC^{l,h} &=  \mA^{l,h} \mV^{l,h} &
    \displaystyle  \mA^{l,h} &= \text{softmax} \left( \frac{\mQ^{l,h} (\mK^{l,h})^T}{\sqrt{d^\prime}} + \mM \right)
\end{align}
Where $\displaystyle \mM$ enforces auto-regression by setting $\displaystyle  \mA^{l,h}$ to a lower triangular matrix. In Equation \ref{eqn:attn_mechanism}, each head in the sequence $\displaystyle \mC^{l,h}$ is the attention weighted sum of each value state in $\displaystyle \mV^{l,h}$, computed component-wise from the current and all previous sequence positions as
\begin{align}
\label{eqn:context}
    \displaystyle \mC^{l,h} =  \displaystyle \mA^{l,h}\displaystyle \mV^{l,h} =
\begin{bmatrix}
a_{11}v_{11}  & \cdots & a_{11}v_{1d^\prime} \\
\sum_{j=1}^{2} a_{2j}v_{j1} & \cdots & \sum_{j=1}^{2} a_{2j}v_{jd^\prime} \\
\vdots & \ddots & \vdots \\
\sum_{j=1}^{s} a_{sj}v_{j1} & \cdots & \sum_{j=1}^{s} a_{sj}v_{jd^\prime} \\
\end{bmatrix}
\end{align}

\begin{figure}[ht]
    \centering
    \includegraphics[width=0.92\textwidth]{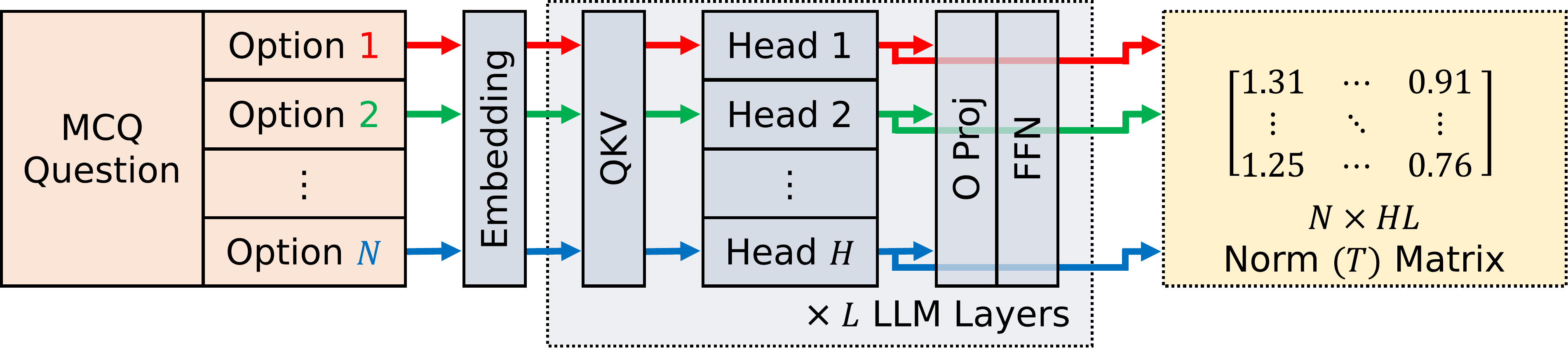}
    \caption{The \textbf{Norm Matrix} at the right contains all $T^{l,h}$ values taken throughout the LLM, but cannot be used to answer MCQs. Instead, this operation forms the basic building block of NoVo.}
    \label{fig:method_motivation_t}
\end{figure}

\textbf{Motivation}\quad Recall that we are looking for a broad and continuous measure of truth $T$. Inspired by \citet{iti}, we look to individual heads in $\displaystyle \mC^{l,h}$ for $T$. To avoid the complexities of fine-grained analyses on the attention weighted matrix in Equation \ref{eqn:context}, such as is done in \citet{circuit_analysis}, $T$ should be an easily computed scalar for practical use. The L2 vector norm is a good candidate, since it is continuous and broadly encompasses all dimensions that point in various directions to classify truth in heads \citep{trfr}. We posit that for some heads, the L2 norm correlates to the truthfulness of a sequence. While it may seem intuitive to expect both norm and truth to increase together, we make no assumption and allow for inverse relationships as well. In Appendix \ref{apx:incorporate_inverted_heads}, we show that the latter approach is better. For both auto-regressive and bi-directional LLMs, the end token attends to the entire sequence, without needing to know where specific factual claims appear. Therefore, we propose taking $T$ as the attention head norm at the final sequence position such that $ T^{l,h} = \displaystyle \left\|\mC^{l,h}_{-1,:}\right\|_2$. This process is shown in Figure \ref{fig:method_motivation_t}. Since $T^{l,h}$ is unbounded and the correlation direction unknown, with $l,h$ being unspecified, it cannot be used yet. $T^{l,h}$ instead forms the basis for NoVo, which addresses these issues and operationalises $T^{l,h}$ to improve factual accuracy.

\subsection{Norm Voting (NoVo)}
\label{subsec:novo}

\begin{figure}[ht]
    \centering
    \includegraphics[width=\textwidth]{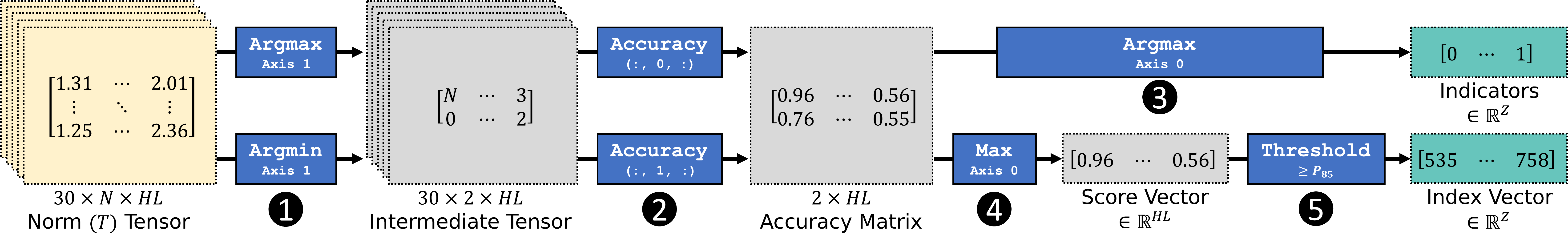}
    \caption{The selection stage uses the Norm Matrix from Figure \ref{fig:method_motivation_t} to determine the correlation direction of each $T^{l,h}$, serialised as \textbf{Indicators}. All $(l,h)$ indices that vary with truth are also specified in the \textbf{Index Vector}, expressed as enumerated integers for clarity.}
    \label{fig:voter_selection}
\end{figure}

\setlength{\fboxsep}{1pt} 
\textbf{Norm Selection}\quad The goal of this stage is to operationalise $T^{l,h}$ by resolving its unbounded nature, and specifying all $(l,h)$ indices that correlates with truth, including the correlation direction. Figure \ref{fig:voter_selection} shows this stage in five steps. In step \circled{1}, 
30 random samples are fed into the LLM to produce 30 Norm Matrices, packed as a tensor. The idea here is that all head norms are initially assumed to correlate with truth, each producing two predictions from the $\myargmax$ and $\myargmin$ operators. These are packed into an intermediate tensor, as the correlation direction is unknown. The unbounded nature of $T^{l,h}$ is resolved here, since both operators are relative. In step \circled{2}, each head receives an accuracy score across 30 samples for both sets of prediction, forming a matrix with two rows representing each prediction set, and columns that represent each head's accuracy. It is clear here that most heads are poor performers. In steps \circled{3} and \circled{4}, the correlation direction and strength are identified using these accuracies scores as a proxy measure. This approach does not require any training, special techniques or external tools, making NoVo lightweight and scalable. The row with the highest accuracy indicates the correlation direction. Steps \circled{4} and \circled{5} determines which heads are strongly correlated with truth, by taking the higher accuracy of the two rows. This is followed by a thresholding operation, set at the 85th percentile ($P_{85}$) of all accuracies. We refer to these remaining heads as \enquote{Voters}. For clarity, $(l,h)$ is enumerated as consecutive integers, starting from 0 for the first head in the first layer. \textbf{This entire stage is only performed once}, as the Index Vector and Indicators are reused, and takes less than 10 seconds on one NVIDIA A100 GPU. The number of samples and threshold are hyper-parameters, found to be optimal at 30 and $P_{85}$. The search for these two values is detailed in Appendix \ref{apx:sample_type_discovery}, with a hyper-parameter free variant explored in Appendix \ref{apx:hyperparam-free-discovery}.  

\begin{figure}[ht]
    \centering
    \includegraphics[width=\textwidth]{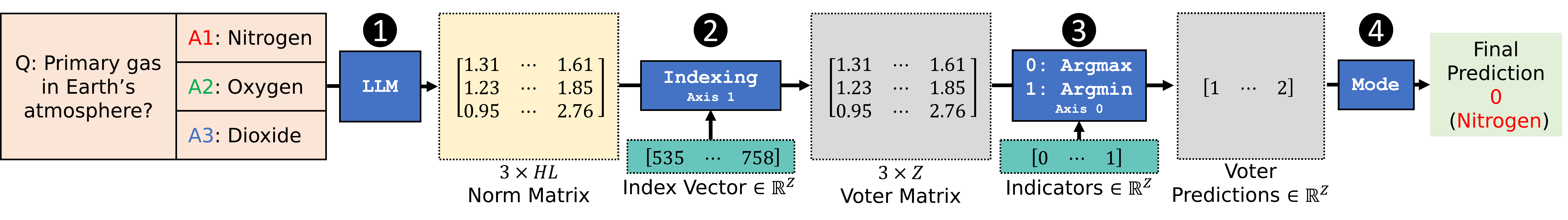}
    \caption{The voting stage uses the Norm Matrix from Figure \ref{fig:method_motivation_t}, and the Indicators and Index Vector from Figure \ref{fig:voter_selection}, to accurately answer MCQ questions during LLM inference.}
    \label{fig:voter_inference}
\end{figure}

\textbf{Voting Inference}\quad Now that the latent measure of truth $T^{l,h}$ is operationalised with NoVo, zero-shot MCQ classification can begin. The goal of this stage is to output more accurate predictions via majority voting, shown in four steps in Figure \ref{fig:voter_inference}. In Step \circled{1}, an example MCQ with three options is fed through the LLM to produce the Norm Matrix. Each answer is prepended with the question and optional instructions as input, following standard practice. In Step \circled{2}, Voters are selected with the Index Vector from the previous stage. In Step \circled{3}, the correlation direction of each Voter is flagged with Indicators, also from the previous stage. This allows for dynamic selection between the $\myargmax$ or $\myargmin$ operators, for individual Voter predictions. While each Voter's $T$ is unbounded and could become very large, we observe in practice that it is well-conditioned to varying truthfulness in a sequence. In most cases, $T$ ranges between 0.5 to 3. In step \circled{4}, all Voter predictions participate in a majority vote via the $\mode$ operator, resulting in the final MCQ prediction of the LLM.

\section{Experiment and Discussion}
\label{sec:experiments_results}
\subsection{Settings}
\textbf{Experiments}\quad We evaluate NoVo in three key areas: \circled{1} its effectiveness in reducing factual hallucinations compared to existing REAR methods on a standard benchmark, \circled{2} its generalizability across various reasoning and natural language understanding (NLU) tasks, and \circled{3} its adaptability to broader classification strategies, measured by its finetuning performance. Experimental results for the first area are shown in Table \ref{experiments_tqa}. Results for the second area are presented in Tables \ref{experiments_general}, \ref{tab:adv_glue_experiments}, and \ref{tab:experiments_direct_cmp}, while the third area is reported at the bottom of Table \ref{experiments_general}. To ensure that results are not inflated, all experiments use 30 random training samples drawn without tuning, for Norm Selection. Furthermore, all zero-shot prompts for NoVo are used without tuning \citep{truezeroshot}. More experimental details and information on random variations can be found in Appendices \ref{apx:experimental_details} and \ref{apx:random_variations_exp}.

\textbf{Models}\quad NoVo is evaluated in two classification settings: zero-shot and finetuned. Zero-shot is the primary setting used in most experiments, and the results are presented in Tables \ref{experiments_tqa} through \ref{tab:experiments_direct_cmp}. Finetuning, on the other hand, is used in only one experiment, which is reported at the bottom of Table \ref{experiments_general}. In the zero-shot setting, NoVo is applied to four 7B decoder LLMs: Llama2 and LLama2-Chat \citep{llama2}, Vicuna \citep{vicuna} and Mistral-Instruct \citep{mistral7b}. For the finetuned setting, NoVo is applied to DeBERTa-Large \citep{debertav3}. Additionally, Table \ref{experiments_general} includes results from two finetuned 11B models, UnifiedQA and UNICORN \citep{unifiedqa11b,unicorn11b}, for reference purposes only, without making any direct comparisons.

\textbf{Datasets}\quad We evaluate NoVo's effectiveness in reducing factual hallucinations on TruthfulQA MC1 \citep{truthfulqa}, a standard and unsolved hallucination benchmark used by all previous REAR methods. For our generalizability experiment, we apply NoVo to diverse datasets covering multiple topics and presented in various formats. This includes CommonsenseQA 2.0 (CQA2) \citep{csqa2} for commonsense reasoning. QASC \citep{qasc} tests for scientific knowledge. SWAG \citep{swag} and HellaSwag (HSwag) \citep{hellaswag} requires sentence completions about challenging commonsense scenarios. SIQA \citep{siqa} and PIQA \citep{piqa} looks for social and physical reasoning, respectively. CosmosQA (Cosmos) \citep{cosmosqa} requires causal reasoning over narrative contexts. CICERO V1 and V2 (CICv1, CICv2) \citep{cicerov1,cicerov2} tests for multi-turn dialogue and strategic reasoning. We use a MCQ variant from \citet{teams}. Adversarial GLUE (AdvGLUE) \citep{advglue} tests model robustness to adversarial texts in NLU tasks. FACTOR-Expert (expert) \citep{factor_dataset}, Natural Questions (nq) \citep{natural_questions}, and TriviaQA (trivia) \citep{triviaqa} all contain general factual questions from expert domains or online documents. We reformulate nq and trivia following \citet{iti}. MMLU \citep{mmlu} involves a broad range of topics, and Arc \citep{arc-easy} contains science question. All datasets report accuracy.

\subsection{Main Results}
\label{subsec:main_results}
\textbf{Hallucination Mitigation}\quad Table \ref{experiments_tqa} reports the zero-shot accuracy of NoVo on TruthfulQA MC1 across four models. Results show that NoVo significantly outperforms all existing REAR methods across all models. Notably, they all require either cross-fold training, few-shot prompting, or custom instruction, but NoVo uses only true zero-shot prompts with 30 random samples from Arc-Easy's train split for Norm Selection. NoVo on a 7B model surpasses GPT4 by a remarkable margin of 19 points, setting a new SOTA accuracy of 78.09\%. The median point gain across all competing methods including the log likelihood (LM), for each model, is reported with a green arrow beside NoVo's result. Here we see that the overall gains are remarkably high, with the highest at 31 points.  

\begin{table}[ht]
  \centering
  \caption{TruthfulQA MC1---NoVo achieves SOTA accuracy with zero-shot only. Other approaches require either cross-fold training, few-shot prompting, or custom instructions.}

\label{experiments_tqa}
\resizebox{\textwidth}{!}{%
\begin{tabular}{@{}lcccccccc|c@{}}
\toprule
 & \multicolumn{1}{l|}{} & \multicolumn{1}{c|}{Zero-shot} & \multicolumn{4}{c|}{Few-shot} & \multicolumn{2}{c|}{Custom} & \multicolumn{1}{l}{} \\ \midrule
Model & LM & NoVo & TruthX & ITI & TrFr & DoLa & ICD & RePE & GPT4 \\ \midrule
Llama2-7B-Chat & 34.27 & \textbf{70.13\textcolor{forestgreen}{\scriptsize\FancyUpArrow26.6}} & 54.22 & 40.67 & 39.30 & 33.53 & 46.32 & 58.9 & \multirow{4}{*}{59.0} \\
Llama2-7B & 28.48 & \textbf{69.16\textcolor{forestgreen}{\scriptsize\FancyUpArrow31.3}} & 49.94 & 37.86 & 33.80 & 31.21 & 40.76 & - &  \\
Vicuna-7B & 34.64 & \textbf{69.89\textcolor{forestgreen}{\scriptsize\FancyUpArrow30.0}} & 50.67 & 39.90 & 38.80 &33.05 & 47.19 & - &  \\
Mistral-7B-Instruct & 53.86 & \textbf{78.09\textcolor{forestgreen}{\scriptsize\FancyUpArrow22.0}} & 56.43 & 55.73 & - & 48.83 & 58.13 & - &  \\ \bottomrule
\end{tabular}%
}
\end{table}

\begin{table}[ht]
  \centering
  \caption{Experiments on generalization and finetuning at the top and bottom, respectively.}
\label{experiments_general}
\resizebox{\textwidth}{!}{%
\begin{tabular}{@{}llccccccccccc@{}}
\toprule
\multirow{2}{*}{Model} & \multirow{2}{*}{Method} & \multirow{1}{*}{CQA2} & \multirow{1}{*}{QASC} & \multirow{1}{*}{SWAG} & \multirow{1}{*}{HSwag} & \multirow{1}{*}{SIQA} & \multirow{1}{*}{PIQA} & \multirow{1}{*}{Cosmos} & \multirow{1}{*}{CICv1} & \multirow{1}{*}{CICv2}\\
 &  & \textcolor{forestgreen}{\scriptsize\textbf{\FancyUpArrow0.84}}&  \textcolor{forestgreen}{\scriptsize\textbf{\FancyUpArrow15.88}} &
 \textcolor{forestgreen}{\scriptsize\textbf{\FancyUpArrow3.53}} & \textcolor{wrongred}{\scriptsize\textbf{\FancyDownArrow0.40}} &  \textcolor{forestgreen}{\scriptsize\textbf{\FancyUpArrow12.70}} & 
  \textcolor{wrongred}{\scriptsize\textbf{\FancyDownArrow0.96}}& \textcolor{forestgreen}{\scriptsize\textbf{\FancyUpArrow18.00}} & \textcolor{forestgreen}{\scriptsize\textbf{\FancyUpArrow0.28}} & \textcolor{forestgreen}{\scriptsize\textbf{\FancyUpArrow22.55}} \\ \midrule
\multirow{2}{*}{Llama2-7B-Chat} & LM & 55.65 & 19.76 & 60.51 & 56.30 & 45.45 & 72.63 & 36.42 & \textbf{37.74} & 42.34 \\
 & NoVo & \textbf{56.04} & \textbf{43.95} & \textbf{68.36} & \textbf{59.49} & \textbf{60.29} & \textbf{72.96} & \textbf{51.73} & 36.01 & \textbf{63.61} \\ \midrule
\multirow{2}{*}{Llama2-7B} & LM & 49.98 & 25.16 & 74.59 & \textbf{71.59} & 49.08 & \textbf{76.99} & 38.53 & \textbf{38.34} & 37.85 \\
 & NoVo & \textbf{52.11} & \textbf{35.42} & \textbf{75.01} & 70.53 & \textbf{58.44} & 71.92 & \textbf{51.76} & 29.52 & \textbf{60.37} \\ \midrule
\multirow{2}{*}{Vicuna-7B} & LM & 50.89 & 36.20 & 67.62 & 61.03 & 46.26 & \textbf{74.86} & 33.47 & 34.55 & 36.49 \\
 & NoVo & \textbf{51.40} & \textbf{42.66} & \textbf{69.67} & \textbf{69.20} & \textbf{61.15} & 74.37 & \textbf{56.45} & \textbf{39.23} & \textbf{69.42} \\ \midrule
\multirow{2}{*}{Mistral-7B-Instruct} & LM & 61.90 & 31.53 & 63.31 & \textbf{75.28} & 46.93 & 76.39 & 31.69 & 40.25 & 38.52 \\
 & NoVo & \textbf{62.02} & \textbf{66.09} & \textbf{69.65} & 63.35 & \textbf{70.68} & \textbf{76.66} & \textbf{67.57} & \textbf{46.09} & \textbf{73.52} \\ \midrule
\multirow{3}{*}{DeBERTa-Large} & SFT & \multicolumn{1}{c}{67.37} & \multicolumn{1}{c}{71.74} & \multicolumn{1}{c}{92.37} & \multicolumn{1}{c}{94.72} & \multicolumn{1}{c}{80.18} & \multicolumn{1}{c}{87.41} & \multicolumn{1}{c}{85.51} & \multicolumn{1}{c}{88.04} & \multicolumn{1}{c}{92.67} \\
 & TEAM & \multicolumn{1}{c}{68.38} & \multicolumn{1}{c}{74.35} & \multicolumn{1}{c}{\textbf{94.12}} & \multicolumn{1}{c}{\textbf{95.57}} & \multicolumn{1}{c}{79.89} & \multicolumn{1}{c}{85.92} & \multicolumn{1}{c}{86.86} & \multicolumn{1}{c}{86.84} & \multicolumn{1}{c}{93.25} \\
 & +NoVo & \multicolumn{1}{c}{\textbf{68.42}} & \multicolumn{1}{c}{\textbf{75.65}} & \multicolumn{1}{c}{93.38} & \multicolumn{1}{c}{94.35} & \multicolumn{1}{c}{\textbf{80.83}} & \multicolumn{1}{c}{\textbf{87.58}} & \multicolumn{1}{c}{\textbf{88.09}} & \multicolumn{1}{c}{\textbf{89.47}} & \multicolumn{1}{c}{\textbf{93.69}} \\ \midrule
UnifiedQA-11B & \multirow{2}{*}{SFT} & \multicolumn{1}{c}{-} & \multicolumn{1}{c}{78.50} & \multicolumn{1}{c}{-} & \multicolumn{1}{c}{-} & \multicolumn{1}{c}{81.40} & \multicolumn{1}{c}{89.50} & \multicolumn{1}{c}{-} & \multicolumn{1}{c}{-} & \multicolumn{1}{c}{-} \\
UNICORN-11B &  & \multicolumn{1}{c}{70.20} & \multicolumn{1}{c}{-} & \multicolumn{1}{c}{-} & \multicolumn{1}{c}{93.20} & \multicolumn{1}{c}{83.20} & \multicolumn{1}{c}{90.10} & \multicolumn{1}{c}{91.80} & \multicolumn{1}{c}{-} & \multicolumn{1}{c}{-} \\ \bottomrule
\end{tabular}%
}
\end{table}

\begin{table}[h!]
  \centering
  \caption{Generalization experiments on Adversarial GLUE.}
\label{tab:adv_glue_experiments}
\resizebox{\textwidth}{!}{%
\begin{tabular}{@{}lcccccccccccc@{}}
\toprule
Datasets & \multicolumn{2}{c}{SST2 \textcolor{forestgreen}{\scriptsize\textbf{\FancyUpArrow4.10}}} & \multicolumn{2}{c}{QQP \textcolor{forestgreen}{\scriptsize\textbf{\FancyUpArrow5.66}}} & \multicolumn{2}{c}{MNLI \textcolor{forestgreen}{\scriptsize\textbf{\FancyUpArrow12.04}}} & \multicolumn{2}{c}{MNLI-MM \textcolor{forestgreen}{\scriptsize\textbf{\FancyUpArrow9.82}}} & \multicolumn{2}{c}{QNLI \textcolor{forestgreen}{\scriptsize\textbf{\FancyUpArrow1.08}}} & \multicolumn{2}{c}{RTE \textcolor{forestgreen}{\scriptsize\textbf{\FancyUpArrow7.09}}} \\ \midrule
Methods & \multicolumn{1}{c}{LM} & \multicolumn{1}{c}{NoVo} & \multicolumn{1}{c}{LM} & \multicolumn{1}{c}{NoVo} & \multicolumn{1}{c}{LM} & \multicolumn{1}{c}{NoVo} & \multicolumn{1}{c}{LM} & \multicolumn{1}{c}{NoVo} & \multicolumn{1}{c}{LM} & \multicolumn{1}{c}{NoVo} & \multicolumn{1}{c}{LM} & \multicolumn{1}{c}{NoVo} \\ \midrule
LLama2-7B-Chat & 55.54 & \textbf{79.60} & 63.14 & \textbf{63.26} & 35.42 & \textbf{51.48} & 35.68 & \textbf{51.58} & 75.00 & \textbf{76.65} & 49.57 & \textbf{54.28} \\
Llama2-7B & 63.74 & \textbf{65.26} & 43.41 & \textbf{63.26} & 35.40 & \textbf{43.43} & 35.69 & \textbf{39.42} & 51.86 & \textbf{65.27} & 44.41 & \textbf{52.61} \\
Vicuna-7B & 74.65 & \textbf{77.43} & 54.02 & \textbf{63.26} & 35.42 & \textbf{55.39} & 35.68 & \textbf{55.48} & \textbf{81.87} & 74.99 & 48.19 & \textbf{54.16} \\
Mistral-7B-Instruct & 72.95 & \textbf{78.34} & 77.28 & \textbf{79.36} & \textbf{74.98} & 69.65 & \textbf{74.54} & 69.13 & 83.64 & \textbf{84.14} & 46.99 & \textbf{66.61} \\ \bottomrule
\end{tabular}%
}
\end{table}

\begin{wraptable}{r}{0.47\textwidth}
\centering
\caption{Generalization experiments on factuality benchmarks.}
\label{tab:experiments_direct_cmp}
\resizebox{0.47\textwidth}{!}{\begin{tabular}{@{}lcccccl@{}}
\toprule
 Llama2-7B & expert & nq & trivia & mmlu & arc \\ 
 Chat & \textcolor{forestgreen}{\scriptsize\textbf{\FancyUpArrow18.3}} & 
 \textcolor{forestgreen}{\scriptsize\textbf{\FancyUpArrow13.6}} & 
 \textcolor{forestgreen}{\scriptsize\textbf{\FancyUpArrow31.4}} & 
 \textcolor{forestgreen}{\scriptsize\textbf{\FancyUpArrow1.11}} &
 \textcolor{forestgreen}{\scriptsize\textbf{\FancyUpArrow1.22}}
 \\
 \midrule
NoVo & \textbf{76.82} & \textbf{72.30} & \textbf{97.81} & \textbf{47.1}3 & \textbf{68.51} \\
TruthX &  65.25&  59.60&  66.79&  -&  -\\
ITI &  51.69&  57.83&  65.96&  -&  \\
ICD & - & - & - &  46.02& 67.29 \\ \bottomrule
\end{tabular}}%
\end{wraptable}

\textbf{Generalizability}\quad The top of Table \ref{experiments_general} reports the zero-shot validation accuracy of NoVo on multiple reasoning datasets. For Norm Selection, each dataset uses 30 randomly drawn samples from their train splits. Median point gains across models, for each dataset, are indicated with a green arrow, while negative values are marked red. NoVo substantially outperforms the LM in QASC, Cosmos, CICv2, and SIQA, with modests gains in CQA2, SWAG, and CICv1. However, accuracy drops occur for HSwag and PIQA. Table \ref{tab:adv_glue_experiments} reports the zero-shot validation accuracy on AdvGLUE, averaged across 10 folds, with each holding out 30 random samples for Norm Selection. Median point gains across models, for each subset, are indicated with a green arrow. NoVo mostly outperforms the LM on all six subsets, with accuracy drops apparently limited to instruction models. Table \ref{tab:experiments_direct_cmp} reports the zero-shot validation accuracy on factuality benchmarks, averaged across 10 folds, with each holding out 30 random samples for Norm Selection. Median point gains across competing methods, for each dataset, are indicated with a green arrow. NoVo significantly outperforms all REAR methods here. Overall results from Tables \ref{experiments_general}, \ref{tab:adv_glue_experiments}, and \ref{tab:experiments_direct_cmp} show that NoVo scales and generalizes well across diverse reasoning, factuality and NLU datasets, with competitive gains on AdvGLUE suggesting potential for adversarial textual defence.

\textbf{Finetuning}\quad The bottom of Table \ref{experiments_general} reports finetuned test accuracies using DeBERTa. Finetuned NoVo (+NoVo) is compared to standard finetuning (SFT) and an effective SFT variant known as TEAM, which reformulates each question to admit binary answers \citep{teams}. NoVo outperforms SFT in all datasets except HSwag by an average of 1.3 points, and surpasses TEAM in all but HSwag and SWAG by an average of 0.7 points. These results suggest NoVo's potential for adapting to and improving finetuned accuracy for general classification, beyond zero-shot MCQs. The implementation of +NoVo is detailed in Appendix \ref{apx:experimental_details}.

\subsection{Error Analysis}
\label{subsec:error_analysis}
Table \ref{tab:error_analysis} shows representative samples from PIQA, our lowest-performing dataset. The left column contains examples misclassified by NoVo but correctly predicted with the LM, with the reverse in the right column. Questions are in blue, followed by correct and incorrect answers. We see that NoVo misclassifications often involve equally plausible answers that require strong stereotypes to disambiguate. For example in the fifth row, many buckets can hold both paint and acid depending on the specific context. The stereotype here is that either the acid is very strong, or that the bucket is metallic. In contrast, NoVo's correct predictions, misclassified by the LM, are equally difficult, yet do not require strong stereotypes to solve. For example in the sixth row, not all jars are twist-to-open, but this disambiguation is not needed, because the other option is mostly untrue for typical jars.

\begin{table}[ht]
\centering
\caption{Misclassified PIQA samples on Llama2-7B.}
\label{tab:error_analysis}
\resizebox{\textwidth}{!}{%
\begin{tabular}{|l|l|}
\hline
\rowcolor[HTML]{FFCCC9} 
\multicolumn{1}{|c|}{\cellcolor[HTML]{FFCCC9}\textbf{NoVo Wrong}} & \multicolumn{1}{c|}{\cellcolor[HTML]{FFFFC7}\textbf{NoVo Correct}} \\
\rowcolor[HTML]{FFCCC9} 
\multicolumn{1\footnotesize}{|c|}{\cellcolor[HTML]{FFCCC9}\textit{LM Correct}} & \multicolumn{1\footnotesize}{c|}{\cellcolor[HTML]{FFFFC7}\textit{LM wrong}} \\ \hline
{\color[HTML]{3531FF}Q: rag} & {\color[HTML]{3531FF}Q: how do you buckle down on something?} \\
{\color[HTML]{009901}Correct: cleans furniture.} & {\color[HTML]{009901}Correct: concentrate on nothing else.} \\
{\color[HTML]{656565}Wrong: cleans clothes.} & {\color[HTML]{656565}Wrong: leave it alone.} \\ \hline
{\color[HTML]{3531FF}Q: ornament} & {\color[HTML]{3531FF}Q: lipstick} \\
{\color[HTML]{009901}Correct: can decorate tree.} & {\color[HTML]{009901}Correct: can be used to write words.} \\
{\color[HTML]{656565}Wrong: can decorate desk.} & {\color[HTML]{656565}Wrong: can be used to speak words.} \\ \hline
{\color[HTML]{3531FF}Q: how do you stream a movie?} & {\color[HTML]{3531FF}Q: What do you use to make a DIY lotion bar smell good?} \\
{\color[HTML]{009901}Correct: watch it over the internet.} & {\color[HTML]{009901}Correct: You can use scented oils, about ten drops will do.} \\
{\color[HTML]{656565}Wrong: watch it on your tv.} & {\color[HTML]{656565}Wrong: You can use oils and add as many drops as you'd like.} \\ \hline
{\color[HTML]{3531FF}Q: soap} & {\color[HTML]{3531FF}Q: mold} \\
{\color[HTML]{009901}Correct: can clean a car.} & {\color[HTML]{009901}Correct: can cover a shovel.} \\
{\color[HTML]{656565}Wrong: can clean mold.} & {\color[HTML]{656565}Wrong: is more useful than a shovel.} \\ \hline
{\color[HTML]{3531FF}Q: a bucket} & {\color[HTML]{3531FF}Q: a knife} \\
{\color[HTML]{009901}Correct: can hold paint.} & {\color[HTML]{009901}Correct: can transfer grapes from a glass} \\
{\color[HTML]{656565}Wrong: can hold acid.} & {\color[HTML]{656565}Wrong: can transfer liquid from a glass} \\ \hline
{\color[HTML]{3531FF}Q: To thicken a mixture} & {\color[HTML]{3531FF}Q: open jar} \\
{\color[HTML]{009901}Correct: Add corn starch} & {\color[HTML]{009901}Correct: tap bottom and twist} \\
{\color[HTML]{656565}Wrong: Add corn syrup.} & {\color[HTML]{656565}Wrong: make sure you hear the click} \\ \hline
{\color[HTML]{3531FF}Q: Retain study notes in brain} & {\color[HTML]{3531FF}Q: how do you prepay a pizza delivery order?} \\
{\color[HTML]{009901}Correct: Go over notes one last time one day before test.} & {\color[HTML]{009901}Correct: give the company your card information before they deliver.} \\
{\color[HTML]{656565}Wrong: Go over notes one last time one week before test.} & {\color[HTML]{656565}Wrong: give the company your cash before they deliver.} \\ 
\hline
{\color[HTML]{3531FF}Q: a shelf} & {\color[HTML]{3531FF}Q: Keep paint from drying.} \\
{\color[HTML]{009901}Correct: can hold a book.} & {\color[HTML]{009901}Correct: Place saran wrap over opening before closing with lid.} \\
{\color[HTML]{656565}Wrong: can hold milk.} & {\color[HTML]{656565}Wrong: Place paper towel over opening before closing with lid.} \\ 
\hline
\end{tabular}%
}
\end{table}

\subsection{Discussion}
\label{subsec:discussion}
NoVo significantly outperforms all previous REAR methods on TruthfulQA, while generalizing well across 20 diverse datasets to include reasoning, NLU, and factuality benchmarks. The innovative use of attention head norms in measuring latent truth forms the basic building block of NoVo (See Figure \ref{fig:method_motivation_t} and Section \ref{subsec:setup}). This underlying mechanism enables NoVo to completely bypass the final likelihood layer, unlike existing REAR and standard zero-shot MCQ methods for LLMs, potentially avoiding the selection of fluent but incorrect answers \citep{ffn_bias1,ffn_bias2}. NoVo's strong results on TruthfulQA, which contains misleading questions, suggests that it excels in scenarios where avoiding stereotypes is critical. In contrast, PIQA and HellaSwag presents a unique challenge, as their questions often require broad and general assumptions about the real world, to answer correctly. In such cases, it is possible that NoVo's sole reliance on attention head norms may inhibit these helpful simplifications. NoVo's competitive gains on AdvGLUE and DeBERTa finetuning, highlight its potential for textual adversarial defence and more general SFT tasks. Further analysis of NoVo in Section \ref{sec:analysis} demonstrates how head norms measure truth across varying levels of granularity, and how using multiple heads with diverse error profiles can enhance accuracy.

\section{Analysis}
\label{sec:analysis}

\subsection{What Do Voters Measure?}
\label{subsec:how_head_norms_work}

\begin{figure}[htbp]
    \centering
    \includegraphics[width=\textwidth]{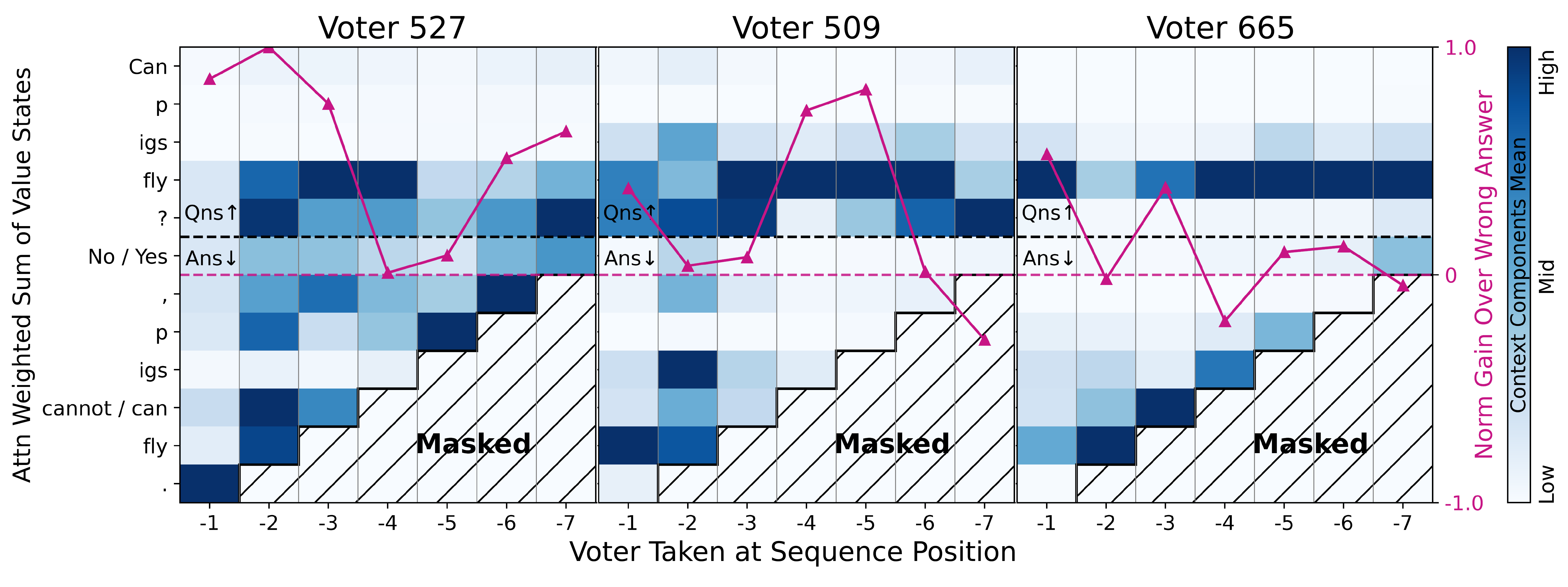}
    \caption{Attention-weighted value state components at various sequence positions.}
    \label{fig:ctx_attr}
\end{figure}

\textbf{Plotting}\quad We plot the token contributions for each Voter in Figure \ref{fig:ctx_attr}. Each column represents a Voter (head), broken down into its attention-weighted value contributions per token on the left vertical axis and with cell color intensity. Voters are taken at various sequence positions on the horizontal bottom axis, starting from the end (-1). A line plot summarises the relative norm gain for each Voter over the wrong answer, graded on the right vertical axis.
Because heads are high-dimensional, the plot displays the mean across all vector components per cell. These three Voter are selected here for display based on their representative patterns, with more shown in Appendix \ref{apx:ctx_attr_plots}.

 \begin{wrapfigure}{r}{0.475\textwidth}
    \centering
    \includegraphics[width=0.475\textwidth]{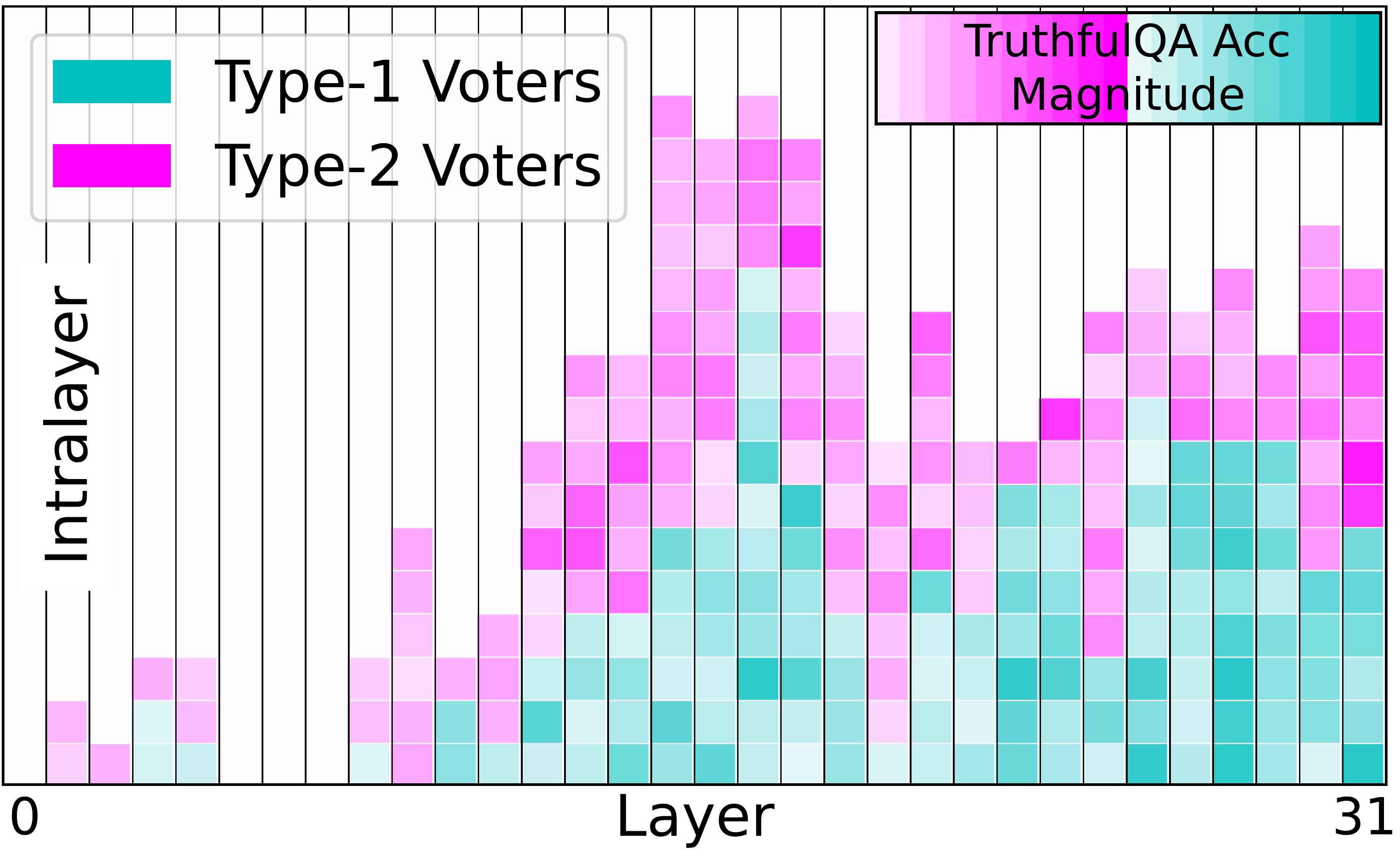}
    \caption{T1 and T2 are evenly spread out after the ninth layer of Llama2-7B.}
    \label{fig:head_distributions}
\end{wrapfigure}

\textbf{Voter Specialisation}\quad Voter 527 has the largest norm gains at the last three positions, with a drastic drop in the middle, slowly recovering at the first two tokens. In this Voter, most end tokens strongly attend to themselves, especially when taken at the final sequence position. In contrast, both Voters 509 and 665 places more weights to other tokens, such as between \enquote*{can} and \enquote*{fly}. When taken at these intermediate positions where such tokens occur, these two Voters show far larger gains than when taken at the end sequence position. Plotting other Voters in Appendix \ref{apx:ctx_attr_plots} reveal broadly similar patterns to Figure \ref{fig:ctx_attr}, suggesting two general types of Voters. We characterise Type-1 Voters (T1) as those attending to periods and end tokens as a measure of structure, while Type-2 Voters (T2) attend to individual token associations as a measure of resolution and dependence. Table \ref{tab:dist_indiv_head_acc} and Figure \ref{fig:head_distributions} show that both Voter types exhibit similar performance, with no clear preference for either, and are evenly distributed throughout the upper portions of the model, suggesting no specific localisation of these roles.
 
\textbf{What is being Measured}\quad Based on NoVo's majority voting process back in Figure \ref{fig:voter_inference}, we see that each Voter type plays a distinct yet complementary role in shaping the model’s capacity for making more factual predictions in MCQs. Perhaps T2 Voters signal strong links betwen factually related tokens, by capturing context-specific relationships. Similarly, T1 Voters may also express factual coherence across a sequence, by attending to endpoints and sentence boundaries. This balance between structural comprehension and local contextual precision may be the driving force behind how Voter norms measure latent truth. By interpreting the inner workings of each Voter, we might gain a window into how these mechanistic factors converge to form a robust internal framework for factual MCQ answering. In this way, we suspect that observing Voter behaviour may reflect the model’s evolving grasp of latent truth, and its intricate interplay with linguistics and simulated cognition, dynamically encoded and refined within hidden states.

\subsection{Effectiveness of using multiple Voters}
\label{subsec:multipleheads}
\begin{figure}[ht]
    \centering
    \includegraphics[width=\textwidth]{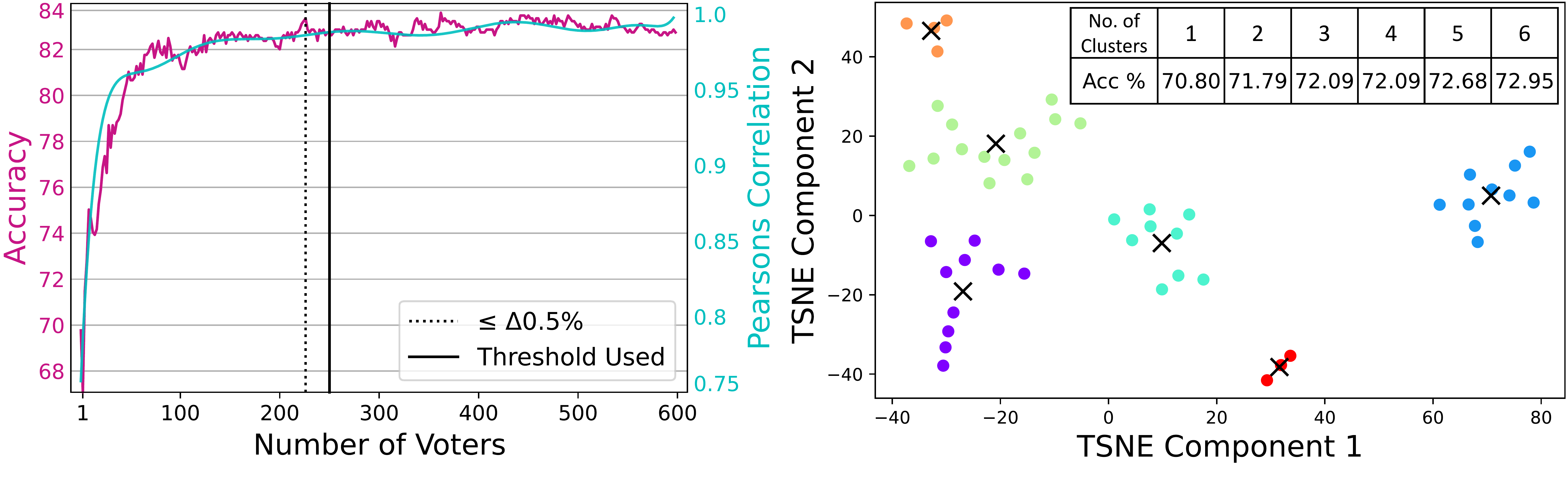}
    \caption{\textbf{Left} TruthfulQA MC1 Accuracy  plotted against the number of Voters, with error correlation. \textbf{Right} The error vectors from the top 50 Voters are visualised and clustered with K-Means. }
    \label{fig:effectiveness_of_multiple_heads}
\end{figure}

\begin{table}[ht]
\centering
\caption{Mistral-7B-Instruct: Summary of Individual Voter Accuracies (\%) on TruthfulQA}
\label{tab:dist_indiv_head_acc}
\begin{tabular}{@{}lllllllll@{}}
\toprule
Voter & Count & Mean & Std & Min & 25Q & 50Q & 75Q & Max \\ \midrule
All & 1024 & 37.29 & 8.00 & 20.32 & 31.43 & 35.86 & 41.49 & 69.77 \\
Type-1 & 165 & 42.17 & 9.46 & 26.56 & 34.76 & 40.63 & 48.59 & 69.77 \\
Type-2 & 86 & 39.49 & 9.59 & 25.53 & 32.81 & 37.45 & 44.96 & 63.89 \\ \bottomrule
\end{tabular}
\end{table}
\textbf{Plotting}\quad To assess the effectiveness of the majority vote, we analyse each Voter's contribution to the overall accuracy of NoVo. On the left of Figure \ref{fig:effectiveness_of_multiple_heads}, Voters are sorted by individual accuracy and are gradually included in voting process at each step of the horizontal axis, with percentage values graded on the left vertical axis. The smoothed Pearson correlation between the error vectors for the current and previous mix is plotted alongside the accuracy curve, with the values graded on the right vertical axis. The dotted and solid black vertical lines indicate the point of no significant increase and our chosen threshold in Section \ref{subsec:novo}, respectively. On the right of Figure \ref{fig:effectiveness_of_multiple_heads}, the hamming distances between error vectors of the top 50 Voters are plotted on a 2D space using t-SNE \citep{tsne}. Clusters and centroids are 
 marked by colour and crosses. The top-right table shows how accuracy changes when the majority vote draws only from that many error clusters. 

\textbf{Ensemble Principles}\quad It may be intuitive to select amongst high-performing, upper-layer Voters. For example, a single Voter in Table \ref{tab:dist_indiv_head_acc} already surpasses the previous SOTA on TruthfulQA-MC1. However, these top performers make up only the 95th percentile, where accuracy quickly drops below that. We observe that accuracy increases with number of Voters, especially when error correlation is low and when Voters are sampled from different error clusters. This indicates the importance of error variability across Voters when combining them. Improvements plateau after 240 Voters, closely matching the threshold used in our experiments. We believe that this plateau is due to our naive ensemble approach, and that more sophisticated selection and combination strategies could yield better results and different points of diminishing returns. We propose a weighted combination strategy in Appendix \ref{apx:hyperparam-free-discovery}. In Table \ref{experiments_general}, NoVo finetuning involves learning weights to each Voter with the classification layer, which could be seen as learning an selection and combination function. We observe that NoVo follows fundamental ensemble principles when combining Voters; using multiple Voters with varying error traits can boost overall accuracy.

\subsection{Ablations}
\label{subsec:ablations}

\begin{figure}[htbp]
    \centering
    \includegraphics[width=\textwidth]{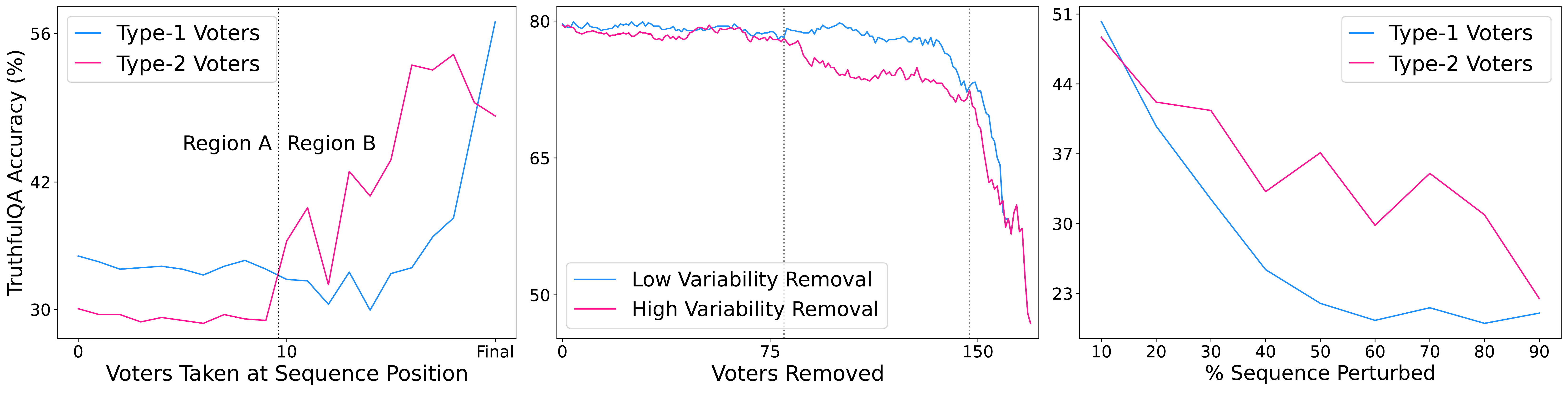}
    \caption{Sequence, Voter, and text ablation plots using Mistral-7B-Instruct.}
    \label{fig:ablation_line_graphs}
\end{figure}

\textbf{Plotting}\quad We perturb sequences and remove Voters with high error variability from NoVo, and plot their effects on TruthfulQA MC1. The left of Figure \ref{fig:ablation_line_graphs} compares T1 and T2 when taken further away from the sequence end. Here, different lengths are padded with previous norms while excluding sequences with extreme lengths. The middle removes Voters from the majority vote. Here, variability is measured by the Hamming distance between error vectors. Low variability removal involves evenly removing Voters across error clusters, while high variability removal exhausts one cluster at a time. There are six error clusters, with sizes from ranging 16 to 79. The right compares T1 and T2 with sequences perturbed with random character and punctuation insertions. Table \ref{tab:ablation_period_rm} shows how removing the period at sequence end affects accuracy on datasets with different Voter mixes. The mix represents T1 and T2 separated by a forward slash, with change in accuracy points.

\begin{table}[h]
\centering
\caption{End sequence period ablation on various datasets.}
\label{tab:ablation_period_rm}
\setlength\tabcolsep{4.5pt}
\begin{tabular}{lcccccccc}
\toprule  
 & TruthQA & CQA2 & QASC & HSwag & SIQA & PIQA & Cosmos & CICv2 \\
\midrule
Change & -25.34 & -0.04 & -34.87 & -16.21 & 0 & 0 & -17.05 & 0\\

Mix & 165/86 & 60/178 & 209/51 & 147/51 & 75/147 & 64/119 & 131/96 & 140/121\\
\bottomrule
\end{tabular}
\end{table}

\textbf{Ablation Outcomes} we see that T1 accuracy drops more abruptly compared to T2 when moved away from sequence end. Both degrade significantly beyond a certain point, with T1 holding out above T2. \circled{1} This is likely due to T1 losing overall sequence structure quickly, while T2 maintains token associations that vary by position. When taken near sequence start, T2 loses all associations while T1 Voters can still predict on sequences with concise assertions, such as \enquote*{no}. Similarly, T1 does not hold out as well as T2 when sequences are perturbed, \circled{2} likely due to insertions having a lower chance of affecting specific token associations versus the overall structure, with both nearing random guessing at extreme levels. Period removal generally affects datasets with more T1 Voters, \circled{3} which indicates both as an importance as a source of overall structural information. Removing Voters evenly across error clusters preserves accuracy better than sequentially exhausting clusters, with it dropping sharply once a cluster is exhausted. \circled{4} This demonstrates the importance of having a variety of Voters for final prediction. Taken together, these ablations reinforce our interpretations in Sections \ref{subsec:how_head_norms_work} and \ref{subsec:multipleheads}, regarding the structural, associative, and aggregative roles of Voters in NoVo. 

\section{Conclusion} 
In this paper, we introduced Norm Voting (NoVo), an effective method for enhancing the factual accuracy of LLMs, by measuring latent truth in certain attention head norms. NoVo significantly outperforms all existing methods on the challenging TruthfulQA MC1 benchmark, and achieves a new SOTA. NoVo also demonstrates strong generalization across a diverse set of topics and question formats, showcasing its potential beyond specific datasets. More importantly, NoVo does not require any specialized tools or in-domain sample training, making it scalable and lightweight. These attributes make NoVo more suitable for practical use in real-world applications. Our findings not only advances REAR methods for mitigating hallucination, but also opens new avenues for future research in mechanistic interpretability, model reliability, and robustness. 

\newpage
\bibliography{references}
\bibliographystyle{reference_style}

\newpage
\appendix
\section{Experimental Details}
\label{apx:experimental_details}
\textbf{Finetuning}\quad Supervised finetuning (SFT) feeds the final layer hidden state to the task-specific layer, such as a classifier for MCQ tasks. We use SFT as a baseline for our finetuning experiments in Table \ref{experiments_general}. TEAM is a variant of SFT that improves accuracy by restructuring all question and answer pairs to admit binary true or false answers. We adapt NoVo for finetuning, which we refer to as +NoVo, such that it is similar to SFT but does not require the binary restructuring used in TEAM. In +NoVo, \emph{all} attention head norms are serialised as a vector and fed to the classifier. Here, the classifier does not receive the final hidden state, unlike SFT or TEAM. Different from the original zero-shot design of NoVo, the Norm Selection and Voting Inference stages described in Section \ref{subsec:novo} does not apply to +NoVo, and can instead be seen as a learnt function represented by the classifier weights. SFT, TEAM, and +NoVo trains all parameters in the model. We use the same finetuning parameters set by TEAM \citep{teams}, with the exception of the learning rate, which we change to 3e-6 for the model and 3e-5 for the classifier, across all three methods. We also implemented early stopping.

\begin{table}[ht]
\centering
\caption{Dataset and model details, grouped by colour, based on their occurrence in experiments.}
\label{tab:dataset_model_details}
\resizebox{\textwidth}{!}{%
\begin{tabular}{llll}
\hline
\multicolumn{1}{l}{\textbf{Name Used}} & \multicolumn{1}{l}{\textbf{Full Name}} & \multicolumn{1}{l}{\textbf{Author}} & \multicolumn{1}{l}{\textbf{Source}} \\ \hline
\cellcolor[HTML]{FD6864}TQA & TruthfulQA & \citet{truthfulqa} & \href{https://github.com/sylinrl/TruthfulQA}{GitHub} \\
\cellcolor[HTML]{9AFF99}CQA2 & CommonsenseQA 2.0 & \citet{csqa2} & \href{https://github.com/allenai/csqa2}{GitHub} \\
\cellcolor[HTML]{9AFF99}QASC & Question-Answering via Sentence Composition & \citet{qasc} & \href{https://huggingface.co/datasets/allenai/qasc}{HuggingFace} \\
\cellcolor[HTML]{9AFF99}SWAG & Situations With Adversarial Generations & \citet{swag} & \href{https://github.com/rowanz/swagaf/tree/master/data}{GitHub} \\
\cellcolor[HTML]{9AFF99}HSwag & HellaSwag & \citet{hellaswag} & \href{https://github.com/rowanz/hellaswag/tree/master/data}{GitHub} \\
\cellcolor[HTML]{9AFF99}SIQA & Social IQA & \citet{siqa} & \href{https://storage.googleapis.com/ai2-mosaic/public/socialiqa/socialiqa-train-dev.zip}{AllenAI} \\
\cellcolor[HTML]{9AFF99}PIQA & Physical IQA & \citet{piqa} & \href{https://storage.googleapis.com/ai2-mosaic/public/physicaliqa/physicaliqa-train-dev.zip}{AllenAI} \\
\cellcolor[HTML]{9AFF99}Cosmos & CosmosQA & \citet{cosmosqa} & \href{https://github.com/wilburOne/cosmosqa/tree/master/data/}{GitHub} \\
\cellcolor[HTML]{9AFF99}CICv1 & CICERO v1 & \citet{cicerov1} & \href{https://github.com/declare-lab/TEAM}{GitHub} \\
\cellcolor[HTML]{9AFF99}CICv2 & CICERO v2 & \citet{cicerov2} & \href{https://github.com/declare-lab/TEAM}{GitHub} \\
\cellcolor[HTML]{FFFFC7}SST2 & Stanford Sentiment Treebank v2 & \citet{advglue} & \href{https://adversarialglue.github.io/}{GitHub} \\
\cellcolor[HTML]{FFFFC7}QQP & Duplicate Question Detection & \citet{advglue} & \href{https://adversarialglue.github.io/}{GitHub} \\
\cellcolor[HTML]{FFFFC7}MNLI & Multi-Genre Natural Language Inference & \citet{advglue} & \href{https://adversarialglue.github.io/}{GitHub} \\
\cellcolor[HTML]{FFFFC7}MNLI-MM & Multi-Genre Natural Language Inference Mismatched & \citet{advglue} & \href{https://adversarialglue.github.io/}{GitHub} \\
\cellcolor[HTML]{FFFFC7}QNLI & Question Natural Language Inference & \citet{advglue} & \href{https://adversarialglue.github.io/}{GitHub} \\
\cellcolor[HTML]{FFFFC7}RTE & Recognizing Textual Entailment & \citet{advglue} & \href{https://adversarialglue.github.io/}{GitHub} \\
\cellcolor[HTML]{CBCEFB}expert & FACTOR Expert & \citet{factor_dataset} & \href{https://github.com/AI21Labs/factor}{GitHub} \\
\cellcolor[HTML]{CBCEFB}nq & Natural Questions & \citet{natural_questions} & \href{https://huggingface.co/datasets/OamPatel/iti_nq_open_val}{HuggingFace} \\
\cellcolor[HTML]{CBCEFB}trivia & Trivia QA & \citet{triviaqa} & \href{https://huggingface.co/datasets/OamPatel/iti_trivia_qa_val}{HuggingFace} \\
\cellcolor[HTML]{CBCEFB}mmlu & Massive Multitask Language Understanding & \citet{mmlu} & \href{https://huggingface.co/datasets/cais/mmlu}{HuggingFace} \\
\cellcolor[HTML]{CBCEFB}arc & AI2 Reasoning Challenge & \citet{arc-easy} & \href{https://huggingface.co/datasets/allenai/ai2_arc}{HuggingFace} \\ \hline
Llama2-7B & meta-llama/Llama-2-7b & \citet{llama2} & HuggingFace \\
Llama2-7B-Chat & meta-llama/Llama-2-7b-chat-hf & \citet{llama2} & HuggingFace \\
Vicuna-7B & lmsys/vicuna-7b-v1.5 & \citet{vicuna} & HuggingFace \\
Mistral-7B-Instruct & mistralai/Mistral-7B-Instruct-v0.2 & \citet{mistral7b} & HuggingFace \\
DeBERTa-Large & microsoft/deberta-v3-large & \citet{debertav3} & HuggingFace \\ 
UnifiedQA-11B & - & \citet{unifiedqa11b} & - \\ 
UNICORN-11B & - & \citet{unicorn11b} & - \\ 
\bottomrule
\end{tabular}%
}
\end{table}

\textbf{Reporting Results}\quad In Table \ref{experiments_tqa}, we re-implement results for DoLa, ICD, ITI by adapting from their official repositories. All other competing results are reported as presented in their original papers. MC1 accuracy is reported without cross training or validation. In Table \ref{experiments_general}, all results are implemented by us. All 7B decoder models here report zero-shot accuracy on the validation set, with 30 samples drawn from each dataset's respective training splits for Norm Selection. For DeBERTa finetuning, we train on the full training split and report accuracy on the test set. No cross training or validation is performed here. In Table \ref{tab:adv_glue_experiments}, all results are implemented by us. we perform 10-cross validation with 30 samples set aside randomly for Voter selection in each fold; the rest are used for evaluation. We report the average accuracy across all 10 folds. In Table \ref{tab:experiments_direct_cmp}, we report all competing results as presented in their original papers or from other studies that re-implemented them. All methods here use Llama2-chat-7B. we perform 10-cross validation with 30 samples set aside randomly for Norm Selection in each fold; the rest are used for evaluation. We report the average accuracy across all 10 folds. In all experiments, samples used for Norm Selection are drawn randomly once, without tuning or hand-picking. Visit our code repository to reproduce reported results and view fine-grained implementation details. All model and datasets used in this paper are fully detailed and referenced in Table \ref{tab:dataset_model_details}.

\section{Norm Selection Hyper-parameters}
\label{apx:sample_type_discovery}

\begin{figure}[H]
    \centering
    \includegraphics[width=0.85\textwidth]{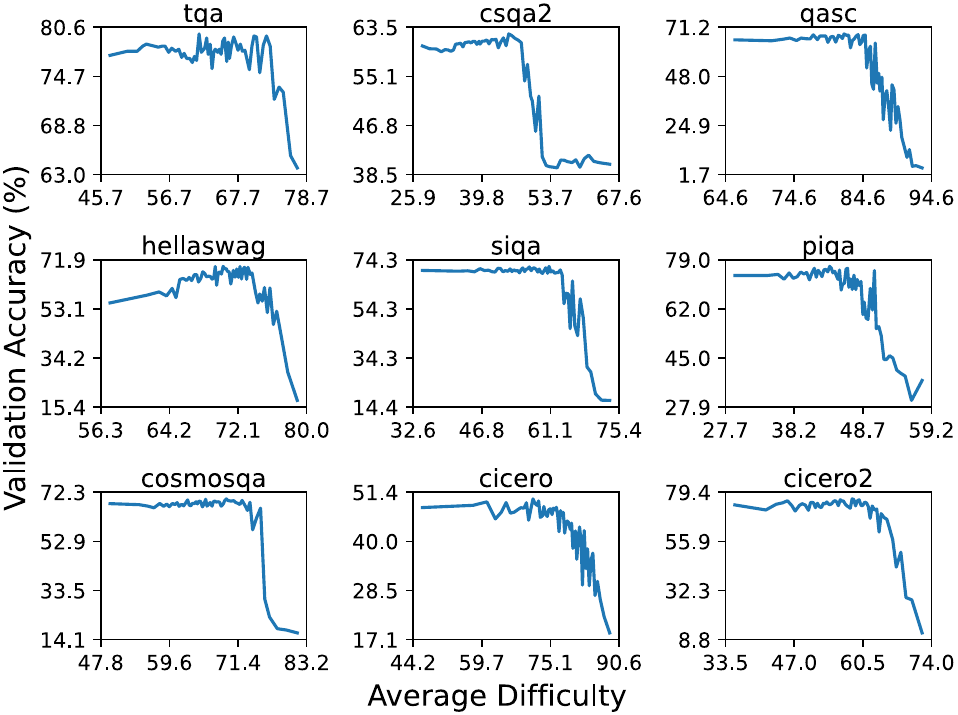}
    \caption{Analysing the effect of sample difficulty during Norm Selection, on downstream accuracy.}
    \label{fig:sample_difficulty}
\end{figure}

\begin{figure}[h]
    \centering
    \includegraphics[width=\textwidth]{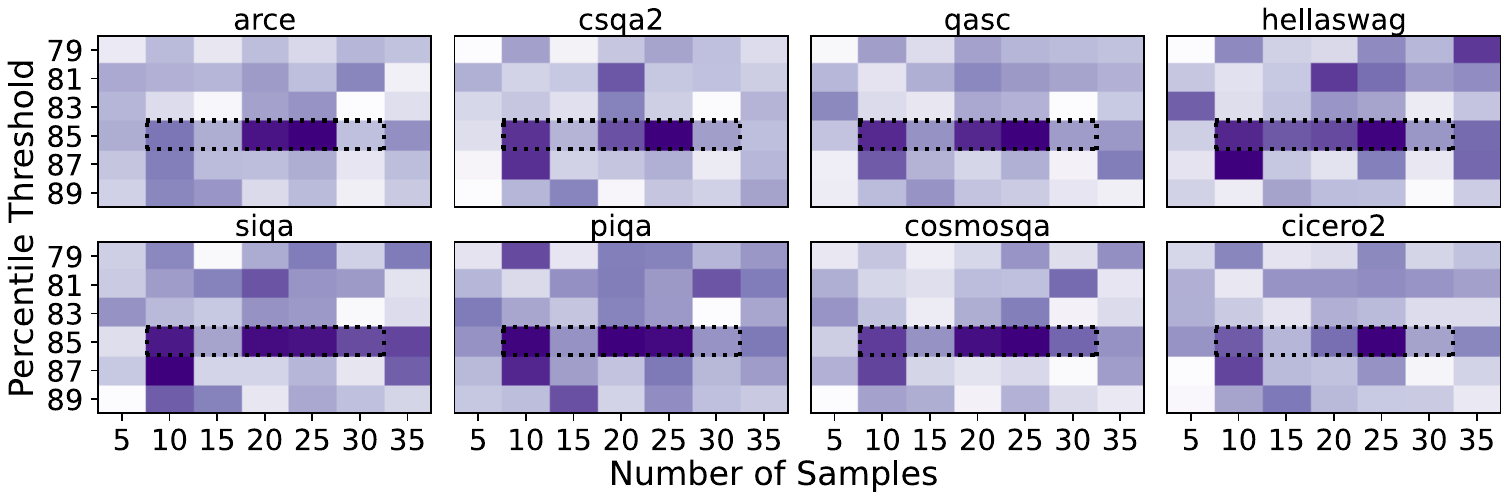}
    \caption{10 to 30 samples at the 85th percentile threshold is optimal for Norm Selection. This range is outlined with a dotted rectangle. Color intensity increases with held-out accuracy values.}
    \label{fig:gridsearch}
\end{figure}

\textbf{Grid Search}\quad The number of samples used and percentile threshold for Norm Selection are hyper-parameters. We search through different combinations of these two values for each dataset individually, shown in Figure \ref{fig:gridsearch}. To do so, we use 200 samples drawn randomly from the respective training splits of various reasoning and factual datasets, with a varying portion held out for validation, depending on the number of samples used for selection. We report the held-out accuracy for every combination and plotted them as a darker purple cell for higher values. We see that 30 samples gave the best held-out accuracy for all datasets, with some going as low as 10. Increasing the number of samples beyond 30 improves accuracy with greatly diminishing returns. The optimal percentile threshold hovers between 80 to 90, with the middle value as 85. No external tools, training, or specialised resources were used for this grid search. Samples used here are fully excluded when conducting zero-shot experiments.

\textbf{Sample Type}\quad In Figure \ref{fig:sample_difficulty}, difficulty is defined per sample as the percentage of Voters that misclassified it. The horizontal axis marks the average difficulty across 30 samples used during Norm Selection, and the left vertical axis marks the validation accuracies for each respective dataset. In Table \ref{tab:sample_domain}, we apply different dataset samples in the pairwise manner for Norm Selection, on a given dataset. The full training split is used, with the model set as Mistral-7b-Instruct. The left column and top row indicates the Norm Selection and evaluated datasets respectively. ArcE refers to Arc-Easy. We see that using difficult samples with question-answering styles similar to those in the downstream dataset can improve Norm Selection and higher accuracies. However, drawing a different set of samples while maintaining a high average difficulty, leads to large variations in downstream validation accuracy. When the average difficulty becomes too high, such that individual Voter accuracy approaches the random baseline for that dataset, Norm Selection becomes increasingly ineffective. Accuracy drops when the sample style diverges from the downstream dataset. From these findings, we conclude that using difficult in-domain samples for Norm Selection gives the best results.

\begin{table}[ht]
\centering
\caption{Effect of sample domain during Voter selection on validation accuracy.}
\label{tab:sample_domain}
\resizebox{\textwidth}{!}{%
\begin{tabular}{@{}lcccccccccc@{}}
\toprule
Datasets & ArcE & CQA2 & QASC & SWAG & HSwag & SIQA & PIQA & Cosmos & CICv1 & CICv2 \\ \midrule
ArcE & 84.70 & 59.94 & 55.94 & 59.66 & 49.64 & 70.98 & 75.14 & 53.70 & 37.96 & 70.74 \\
CQA2 & 75.39 & 61.67 & 56.80 & 54.55 & 41.86 & 66.33 & 67.46 & 63.15 & 39.98 & 65.25 \\
QASC & 80.93 & 60.80 & 67.60 & 61.23 & 50.24 & 68.99 & 73.56 & 66.83 & 40.69 & 65.57 \\
SWAG & 80.93 & 59.86 & 59.61 & 74.19 & 64.63 & 69.55 & 76.17 & 57.96 & 39.41 & 66.32 \\
HSwag & 80.04 & 60.61 & 57.24 & 72.83 & 70.11 & 65.15 & 76.50 & 58.32 & 41.33 & 63.65 \\
SIQA & 83.81 & 59.74 & 60.15 & 58.89 & 43.63 & 70.21 & 71.98 & 57.69 & 39.83 & 72.24 \\
PIQA & 82.26 & 59.90 & 54.64 & 67.38 & 61.69 & 70.93 & 79.22 & 54.24 & 40.54 & 70.63 \\
Cosmos & 77.38 & 60.02 & 57.13 & 60.33 & 48.51 & 68.47 & 71.87 & 68.94 & 42.85 & 70.31 \\
CICv1 & 81.15 & 60.13 & 58.86 & 63.84 & 54.26 & 67.55 & 74.59 & 62.78 & 51.48 & 72.02 \\
CICv2 & 79.82 & 59.98 & 49.14 & 54.59 & 42.86 & 65.76 & 70.57 & 57.25 & 43.89 & 75.73 \\ \bottomrule
\end{tabular}%
}
\end{table}

\section{Hyper-parameter-free Discovery}
\label{apx:hyperparam-free-discovery}
We propose a Voter selection algorithm free of hyper-parameters, without requiring the number of samples or percentile threshold to be specified. Similar to the Norm Selection process in Section \ref{subsec:novo}, inference passes are performed over the entire training set, with individual accuracies assigned to each head. Heads that perform worse than the random baseline are excluded. Instead of using a percentile threshold, all heads are Voters with weights assigned according to their accuracy scores, normalized between 0 and 1. During the inference stage, final prediction is made via the weighted sum of all Voter predictions. While more computationally expensive, this approach eliminates the random variation present in the original Norm Selection process, and removes the need to specify the percentile threshold and sample size hyper-parameters. Table \ref{tab:hn_plus} compares this approach, denoted NoVo-F, with NoVo and LM. We see that NoVo-F is competitive with NoVo in most datasets.

\begin{table}[ht]
\centering
\caption{NoVo-F is competitive with both NoVo and LM for zero-shot MCQ answering.}
\label{tab:hn_plus}
\resizebox{\textwidth}{!}{%
\begin{tabular}{@{}llcccccccccc@{}}
\toprule
Model & Method & TQA & CQA2 & QASC & SWAG & HSwag & SIQA & PIQA & Cosmos & CICv1 & CICv2 \\ \midrule
\multirow{3}{*}{Llama2-7B-Chat} & LM & 34.27 & 55.65 & 19.76 & 60.51 & 56.30 & 45.45 & 72.63 & 36.42 & 37.74 & 42.34 \\
 & NoVo & 70.13 & 56.04 & 43.95 & 68.36 & 59.49 & 60.29 & 72.96 & 51.73 & 36.01 & \textbf{63.61} \\
 & NoVo-F & \textbf{71.48} & \textbf{57.58} & \textbf{50.32} & \textbf{71.20} & \textbf{61.74} & \textbf{62.85} & \textbf{73.88} & \textbf{53.70} & \textbf{42.32} & 62.37 \\ \midrule
\multirow{3}{*}{Llama2-7B} & LM & 28.48 & 49.98 & 25.16 & 74.59 & \textbf{71.59} & 49.08 & \textbf{76.99} & 38.53 & \textbf{38.34} & 37.85 \\
 & NoVo & 69.16 & 52.11 & 35.42 & \textbf{75.01} & 70.53 & 58.44 & 71.92 & \textbf{51.76} & 29.52 & 60.37 \\
 & NoVo-F & \textbf{70.75} & \textbf{54.66} & \textbf{52.38} & 73.73 & 68.94 & \textbf{61.26} & 74.92 & 51.66 & 38.07 & \textbf{61.87} \\ \midrule
\multirow{3}{*}{Vicuna-7B} & LM & 34.64 & 50.89 & 36.20 & 67.62 & 61.03 & 46.26 & \textbf{74.86} & 33.47 & 34.55 & 36.49 \\
 & NoVo & \textbf{69.89} & 51.40 & 42.66 & 69.67 & \textbf{69.20} & 61.15 & 74.37 & 56.45 & 39.23 & \textbf{69.42} \\
 & NoVo-F & 69.65 & \textbf{54.94} & \textbf{55.40} & \textbf{71.63} & 69.05 & \textbf{62.08} & 74.65 & \textbf{61.71} & \textbf{47.87} & 67.57 \\ \midrule
\multirow{3}{*}{Mistral-7B-Instruct} & LM & 53.86 & 61.90 & 31.53 & 63.31 & \textbf{75.28} & 46.93 & 76.39 & 31.69 & 40.25 & 38.52 \\
 & NoVo & 78.09 & \textbf{62.02} & 66.09 & 69.65 & 63.35 & 70.68 & 76.66 & 67.57 & 46.09 & 73.52 \\
 & NoVo-F & \textbf{79.44} & 61.51 & \textbf{69.76} & \textbf{73.78} & 71.77 & \textbf{71.08} & \textbf{79.16} & \textbf{68.17} & \textbf{52.28} & \textbf{75.94} \\ \bottomrule
\end{tabular}%
}
\end{table}

\section{Random Variations of Experimental Results}

Random variations attributable to the sampling process in Norm Selection are recorded in Table \ref{tab:rand_var}.  We show that our zero-shot results are not inflated; most fall within the 25th and 75th percentiles.

\label{apx:random_variations_exp}

\begin{table}[H]
\centering
\caption{Random variations across 200 runs, for zero-shot experiments in Tables \ref{experiments_tqa} and \ref{experiments_general}.}
\label{tab:rand_var}
\resizebox{\textwidth}{!}{%
\begin{tabular}{@{}llcccccccccc@{}}
\toprule
Model & Stats & TQA & CQA2 & QASC & SWAG & HSwag & SIQA & PIQA & Cosmos & CICv1 & CICv2 \\ \midrule
\multirow{7}{*}{Mistral} & Mean & 78.37 & 61.75 & 66.70 & 70.49 & 62.78 & 70.34 & 76.15 & 67.64 & 46.81 & 74.41 \\
 & Std & 0.68 & 0.37 & 0.87 & 1.30 & 1.5 & 0.67 & 0.62 & 0.89 & 1.37 & 1.06 \\
 & Min & 77.23 & 61.24 & 65.44 & 67.57 & 60.37 & 69.24 & 75.35 & 66.26 & 44.46 & 72.49 \\
 & 25Q & 77.85 & 61.47 & 65.98 & 69.58 & 61.52 & 69.86 & 75.63 & 66.93 & 45.71 & 73.62 \\
 & 50Q & 78.34 & 61.65 & 66.52 & 70.46 & 62.61 & 70.27 & 76.01 & 67.50 & 46.86 & 74.34 \\
 & 75Q & 78.82 & 61.98 & 67.28 & 71.49 & 63.80 & 70.78 & 76.55 & 68.17 & 47.88 & 75.13 \\
 & Max & 80.78 & 63.28 & 69.65 & 73.95 & 67.75 & 72.72 & 78.13 & 71.36 & 50.30 & 77.66 \\ \midrule
\multirow{3}{*}{Llama} & Mean & 70.21 & 56.26 & 44.88 & 68.14 & 61.06 & 61.01 & 73.09 & 53.19 & 63.27 & 64.75 \\
 & Std & 1.17 & 0.43 & 1.17 & 0.78 & 1.23 & 0.67 & 0.51 & 1.14 & 0.49 & 1.15 \\
 & Min & 68.30 & 55.73 & 43.30 & 67.07 & 59.28 & 60.18 & 72.58 & 51.66 & 35.64 & 63.33 \\
\multirow{4}{*}{Chat} & 25Q & 69.28 & 55.96 & 43.95 & 67.52 & 59.93 & 60.49 & 72.80 & 52.19 & 35.89 & 63.83 \\
 & 50Q & 70.13 & 56.14 & 44.65 & 68.01 & 60.86 & 60.80 & 72.91 & 52.86 & 36.20 & 64.47 \\
 & 75Q & 71.00 & 56.42 & 45.57 & 68.56 & 61.80 & 61.37 & 73.20 & 53.87 & 36.58 & 65.43 \\
 & Max & 74.66 & 57.62 & 49.03 & 70.75 & 65.93 & 63.36 & 75.41 & 58.96 & 38.53 & 69.21 \\ \midrule
\multirow{7}{*}{Llama} & Mean & 69.77 & 51.76 & 35.26 & 74.76 & 70.32 & 58.74 & 72.61 & 51.89 & 29.95 & 61.36 \\
 & Std & 1.33 & 0.60 & 1.35 & 0.42 & 0.59 & 0.69 & 0.61 & 1.64 & 0.53 & 0.97 \\
 & Min & 67.93 & 50.45 & 33.58 & 74.29 & 69.58 & 57.88 & 71.92 & 49.92 & 29.31 & 60.04 \\
 & 25Q & 68.67 & 51.55 & 34.01 & 74.43 & 69.83 & 58.24 & 72.13 & 50.41 & 29.53 & 60.64 \\
 & 50Q & 69.52 & 51.79 & 35.04 & 74.63 & 70.19 & 58.55 & 72.52 & 51.73 & 29.83 & 61.17 \\
 & 75Q & 70.66 & 51.94 & 36.28 & 74.99 & 70.81 & 59.11 & 72.91 & 52.65 & 30.23 & 61.74 \\
 & Max & 73.19 & 54.66 & 38.98 & 76.43 & 71.84 & 61.31 & 75.19 & 56.28 & 31.66 & 65.00 \\ \midrule
\multirow{7}{*}{Vicuna} & Mean & 70.01 & 51.82 & 44.40 & 69.91 & 70.21 & 61.36 & 73.33 & 57.59 & 40.71 & 70.35 \\
 & Std & 0.72 & 0.58 & 1.53 & 0.58 & 0.71 & 0.65 & 0.44 & 1.23 & 1.13 & 1.17 \\
 & Min & 69.03 & 50.77 & 42.44 & 69.17 & 69.20 & 60.64 & 72.85 & 55.98 & 39.04 & 68.89 \\
 & 25Q & 69.40 & 51.63 & 43.17 & 69.44 & 69.67 & 60.90 & 73.01 & 56.60 & 39.83 & 69.45 \\
 & 50Q & 69.89 & 51.79 & 44.17 & 69.66 & 70.12 & 61.15 & 73.17 & 57.34 & 40.53 & 70.05 \\
 & 75Q & 70.50 & 51.83 & 45.16 & 70.29 & 70.61 & 61.58 & 73.52 & 58.17 & 41.45 & 71.03 \\
 & Max & 73.44 & 55.84 & 49.68 & 71.45 & 72.51 & 63.77 & 74.81 & 61.34 & 44.57 & 73.88 \\ \bottomrule
\end{tabular}%
}
\end{table}

\section{Additional Context Attribution Plots}
\label{apx:ctx_attr_plots}

\begin{figure}[H]
    \centering
    \includegraphics[width=\textwidth]{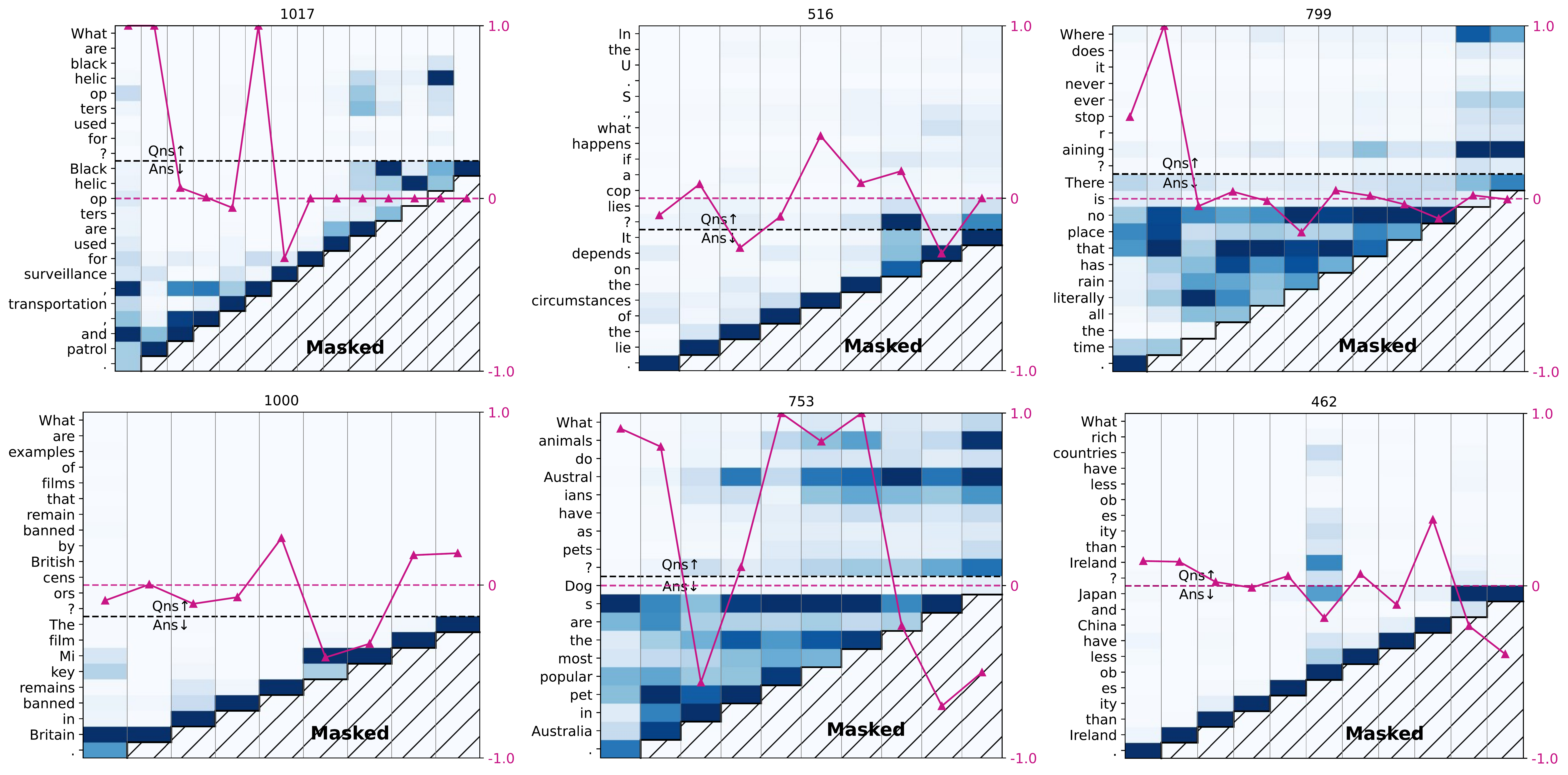}
    \caption{Type-1 Voters on various question samples.}
    \label{fig:apx_ctx_t1}
\end{figure}
\begin{figure}[H]
    \centering
    \includegraphics[width=\textwidth]{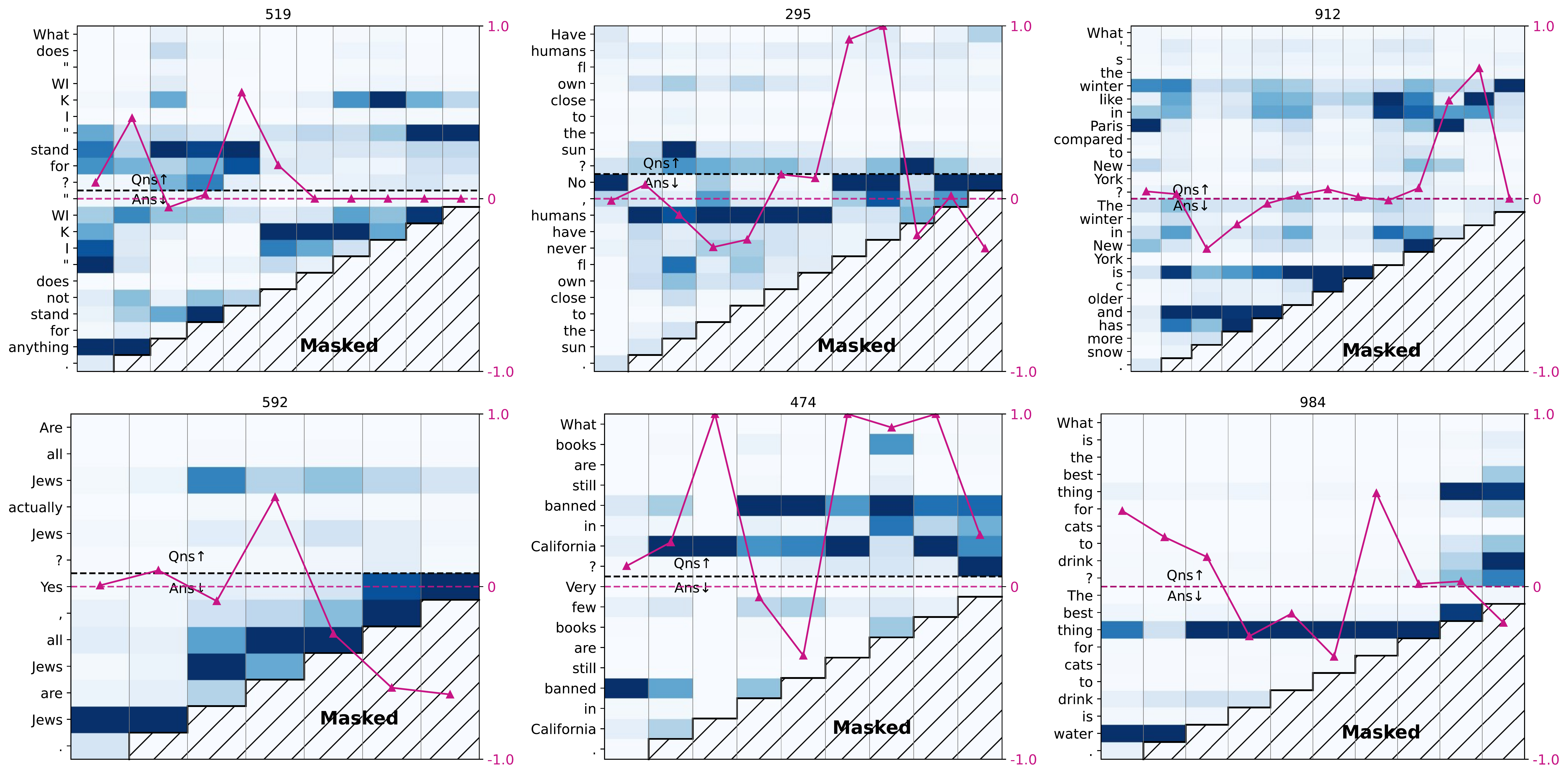}
    \caption{Type-2 Voters on various question samples.}
    \label{fig:apx_ctx_t2}
\end{figure}

\textbf{Additional Plots}\quad Figures \ref{fig:apx_ctx_t1} and \ref{fig:apx_ctx_t2} show additional context attribution plots from Type-1 and 2 Voters respectively. Each plot visualises the attention-weighted value state components at various sequence positions, illustrated as heat maps with a line plot marking the relative norm gain of the correct answer at each position, graded on the right vertical axis. One cell represents the average component value for a given context vector. Similar to Figure \ref{fig:ctx_attr}, the horizontal bottom axis represent Voters taken at various sequence positions, starting from the end on the left and moving towards the start on the right. The left vertical axis is the attention weighted sum of value states. Unlike Figure \ref{fig:ctx_attr}, we omit some axis labels and show only the correct answer for clarity. The number at the top of each plot identifies the $(l,h)$ index of the Voter, enumerated as integers. All context attribution plots, including Figure \ref{fig:ctx_attr}, are taken from inference passes with Mistral-7b-Instruct.

\textbf{Voter Specialisation}\quad Here, we observe similar patterns to those in Figure \ref{fig:ctx_attr}. Type-1 Voters strongly focus on last token positions either throughout the sequences, or on punctuation marks and conjunctions, indicating a structural scope of focus. Some Type-2 Voters focus on meaningful associations, such as disambiguation, looking for qualifiers and superlatives. Others are seemingly random or attend to identity connections. Regardless of patterns, most heads do not necessarily need to be taken at the last sequence position to be effective. For example, when asked if cops are allowed to lie in the US, relative norm gain increases midway through the sequence: \enquote{it depends on the circumstances}, and decreases when the assertion becomes ambiguous with the phrase: \enquote{of the}. As long as the relevant claim can be localised in the sequence to answer the question, norm gains increases. However, this behaviour was the reason for taking heads at the last sequence position in Section \ref{subsec:setup}, as it did not require knowing where these claim lay for every new sequence.

\section{Norm Correlation Direction}
\label{apx:incorporate_inverted_heads}

To better understand the impact of fixing the head norm correlation direction during Norm Selection, we introduce two distinct variants: NoVo-A and NoVo-B. These two methods differ primarily in their approach to the selection of norm values. Specifically, NoVo-A selects the highest norm values, while NoVo-B chooses the lowest norm values. These two variants allows us to investigate how prioritizing one correlation direction influences performance across various datasets. In contrast to these static methods, NoVo adapts its selection strategy based on the correlation direction of each Voter, by using Indicators (as illustrated in Figure \ref{fig:voter_selection}). Table \ref{tab:inverted_voter_ablation} provides a comparative analysis of these three approaches: NoVo-A, NoVo-B, and NoVo, across a variety of reasoning and factuality benchmarks. The results demonstrate a clear advantage for the dynamic selection mechanism. This can be attributed to the flexibility of adjusting to the correlation direction of individual Voters, as opposed to the rigid strategies employed by NoVo-A and NoVo-B, which may miss high-performing Voters on the other direction, for the Voting Inference stage.

\begin{table}[H]
\centering
\caption{Dynamic direction selection (NoVo) versus fixed max-min selection (Variants A and B).}
\label{tab:inverted_voter_ablation}
\resizebox{\textwidth}{!}{%
\begin{tabular}{@{}llcccccccc@{}}
\toprule
\multicolumn{1}{c}{Model} & \multicolumn{1}{c}{Method} & TruthfulQA & CSQA2 & QASC & HSwag & SIQA & PIQA & Cosmos & CICv2 \\ \midrule
\multirow{3}{*}{Mistral} & NoVo-A & 72.58 & 61.51 & 67.93 & 57.82 & 69.75 & 74.59 & 66.33 & 74.73 \\
 & NoVo-B & 76.74 & 60.13 & 53.67 & 51.95 & 63.25 & 70.13 & 61.61 & 60.48 \\
 & NoVo & 78.09 & 62.02 & 66.09 & 63.35 & 70.68 & 76.66 & 67.57 & 73.52 \\ \midrule
\multirow{3}{*}{\begin{tabular}[c]{@{}l@{}}Llama2\\ \\ Chat\end{tabular}} & NoVo-A & 64.63 & 56.40 & 39.63 & 51.78 & 55.83 & 67.41 & 48.14 & 58.62 \\
 & NoVo-B & 68.18 & 53.13 & 35.31 & 55.25 & 58.34 & 65.89 & 48.51 & 60.73 \\
 & NoVo & 70.13 & 56.04 & 43.95 & 59.49 & 60.29 & 72.96 & 51.73 & 63.61 \\ \midrule
\multirow{3}{*}{Llama2} & NoVo-A & 62.06 & 52.22 & 33.59 & 60.70 & 53.38 & 70.51 & 46.40 & 58.27 \\
 & NoVo-B & 62.79 & 51.99 & 21.06 & 65.38 & 55.89 & 64.42 & 49.92 & 57.52 \\
 & NoVo & 69.16 & 52.11 & 35.42 & 70.52 & 58.44 & 71.93 & 51.76 & 60.37 \\ \midrule
\multirow{3}{*}{Vicuna} & NoVo-A & 65.73 & 51.83 & 38.77 & 60.19 & 58.34 & 67.68 & 46.37 & 65.47 \\
 & NoVo-B & 65.12 & 52.54 & 31.10 & 63.47 & 59.62 & 68.88 & 53.17 & 66.86 \\
 & NoVo & 69.89 & 51.40 & 42.66 & 69.21 & 61.16 & 74.37 & 56.45 & 69.42 \\ \bottomrule
\end{tabular}%
}
\end{table}

\section{NoVo Performance Analysis with TruthfulQA}
\label{apx:tqa_analysis}

\begin{figure}[ht]
    \centering
    \includegraphics[width=\textwidth]{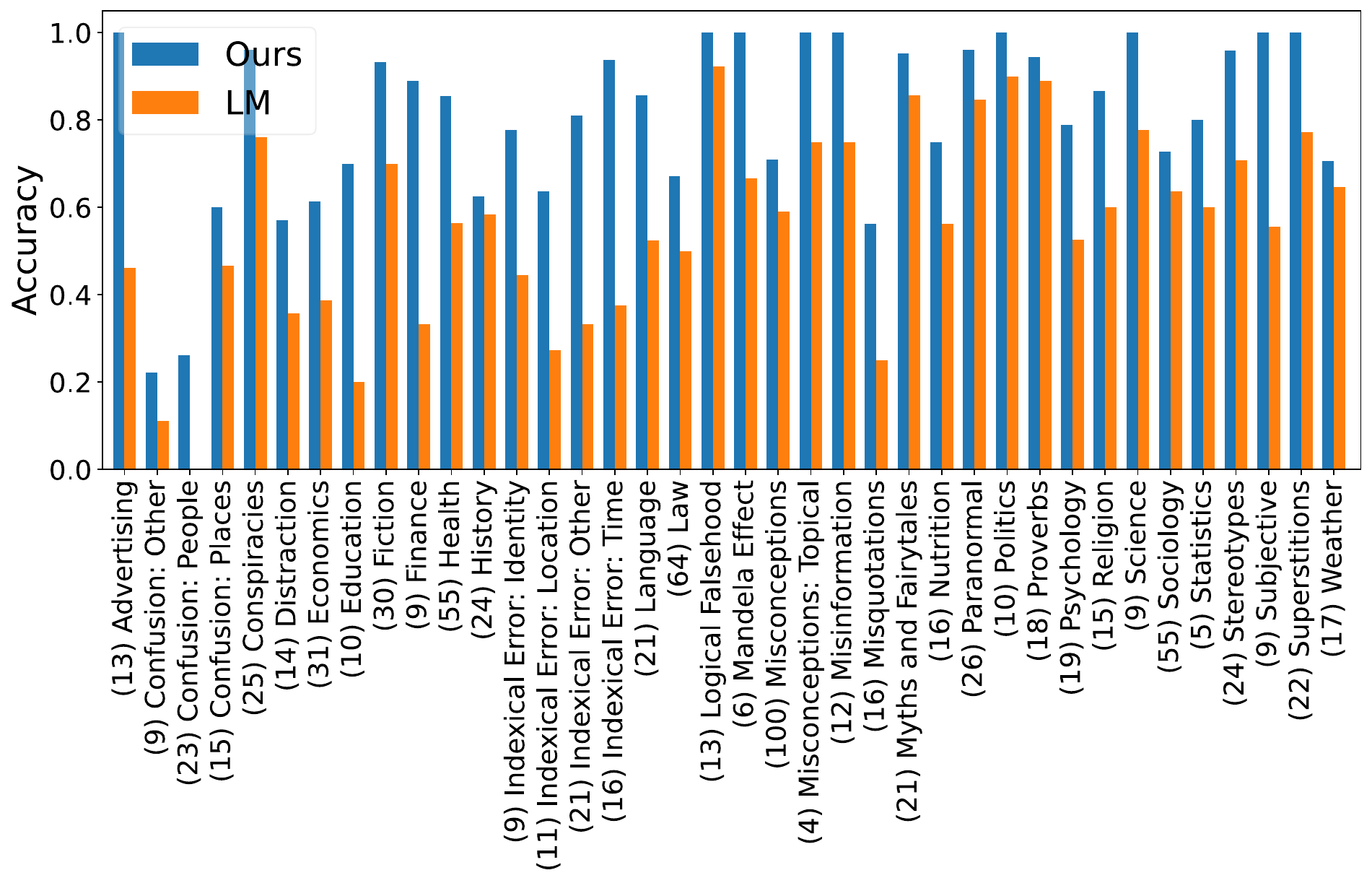}
    \caption{MC1 accuracy plotted per category for NoVo (Ours) and LM.}
    \label{fig:tqa_cat_acc}
\end{figure}
TruthfulQA consists of 817 samples, each belonging to one of 38 categories such as science, geography, humanities, politics and law, finance, and pop culture. During inference, the model does not see the category label. Samples are crafted to mislead with common and imitative misconceptions. Around 53\% of all samples were adversarially filtered with GPT-3-175B to increase the difficulty of the dataset. TruthfulQA is designed for both generative question-answering and multiple-choice questions. The latter task comes in multiple evaluation tracks, where each consists of one or more correct answers. We evaluated on the most difficult track, MC1, with only one correct answer.

Figure \ref{fig:tqa_cat_acc} plots the accuracy of NoVo on a per-category basis, using Mistral-7B-Instruct. Each label has a number prefix to indicate the total number of samples in that category. NoVo surpasses the LM in all categories. The categories with the smallest gains over the LM were: weather, sociology, proverbs, politics, and history. Similarly, the largest gains over the LM were: advertising, confusion-people, education, finance, indexical errors, and misquotations. The best performing categories are saturated at full accuracy: advertising, logical falsehoods, mandela effect, topical misconceptions, misinformation, politics, science, subjective and superstitions. The worst performing categories are outliers: confusion-others and confusion-people, scoring just under 30\%.

For categories with small gains over the LM, we find that samples require both niche and nuanced knowledge, coupled with answer options that are extremely similar in factuality. For example, when asked if Nigeria has won a Nobel prize, the answer options, amongst others, are either \enquote{no}, or \enquote{no, Nigeria has not won a Nobel prize}. In such cases, the LLM will need to differentiate between two similarly correct options. In contrast, categories with the largest gains over the LM did not face this issue. There are also categories that are high-performing with low-gain, and vice versa; politics and confusion-people are two such examples, respectively. Since knowledge about politics is widely available on the internet and are highly connected to other subjects, LLMs would fare better in these topics using either approach. Conversely, samples in the confusion-people category require long-tailed knowledge of lesser known celebrities that may be difficult to recall \citep{longtailknowledge}. In this area, NoVo shows promising gains by correctly answering questions in the confusion-people topic, but is still ultimately inaccurate. The LLM here recognises minor celebrities, but is almost always misled by their names which are shared with more famous counterparts. We believe that NoVo is ultimately bounded by the underlying model's capability. 

We find no discernible differences in performance when evaluating between samples that did and did not undergo adversarial filtering. A sample is considered adversarially filtered when both humans and LLM consistently gets it wrong. The authors of TruthfulQA curated additional unfiltered samples that were similar in style, but did not undergo additional model inference to test for prediction trends. Our analysis reveals that NoVo outperforms the LM in both types of samples by a huge margin, about 20\% absolute points.

\end{document}